\documentclass{article}

\usepackage{iclr2027_conference,times}

\usepackage[utf8]{inputenc}
\usepackage[T1]{fontenc}
\usepackage[hypertexnames=false]{hyperref}
\usepackage{url}
\usepackage{booktabs}
\usepackage{nicefrac}
\usepackage{microtype}
\usepackage{xcolor}
\usepackage{inconsolata}
\usepackage{multirow}
\usepackage{graphicx}
\usepackage{amsmath}
\usepackage{amssymb}
\usepackage{amsfonts}
\usepackage{colortbl}
\usepackage{longtable}
\usepackage{tabularx}
\usepackage{float}
\usepackage{subcaption}
\usepackage{capt-of}
\usepackage{verbatim}
\usepackage{wrapfig}

\definecolor{lgmicolor}{HTML}{0B6E4F}
\definecolor{dnacolor}{HTML}{3B82F6}
\definecolor{irmcolor}{HTML}{EF4444}
\definecolor{binoccolor}{HTML}{D97706}
\definecolor{fdgcolor}{HTML}{7C3AED}
\definecolor{logrankcolor}{HTML}{4B5563}
\definecolor{baseblue}{HTML}{2563EB}
\definecolor{instorange}{HTML}{EA580C}
\definecolor{rolegray}{HTML}{6B7280}
\definecolor{clauseacolor}{HTML}{0EA5E9}
\definecolor{clausebcolor}{HTML}{10B981}
\definecolor{methodgroupbg}{HTML}{F2EAF8}
\newcommand{\methodgrouprow}[2]{\rowcolor{methodgroupbg}\multicolumn{#1}{c}{\textbf{#2}} \\}

\title{Once a Response, Always a Response: Detecting LLM-generated Text via Latent Prompt Restoration}
\hypersetup{%
  pdftitle={Once a Response, Always a Response: Detecting LLM-generated Text via Latent Prompt Restoration},
  pdfauthor={Hongrui Bao, Yubing Ren, Jinhan You, Fang Fang, Shi Wang, Yanan Cao}
}

\author{Hongrui Bao$^{1}$, Yubing Ren$^{2}$, Jinhan You$^{3}$,\\
\textbf{Fang Fang$^{2}$, Shi Wang$^{4}$, Yanan Cao$^{2}$}\\[0.6em]
\normalfont\small $^{1}$University of the Chinese Academy of Sciences\\
\normalfont\small $^{2}$Institute of Information Engineering\\
\normalfont\small $^{3}$Shandong University\\
\normalfont\small $^{4}$Visual Information Processing and Learning}
\date{}

\iclrfinalcopy
\begin{document}

\maketitle
\lhead{}
\begin{abstract}
Large language models (LLMs) can generate fluent and convincing text at scale, creating growing risks for misinformation dissemination, educational misuse, and platform governance. These concerns make robust detection of machine-generated text increasingly necessary. Recent zero-shot detectors mainly exploit probability-based statistical discrepancies, but they do not explicitly model the response-conditioned generation process introduced by instruction tuning. As a result, the latent dependency of machine-generated text on its upstream prompt remains largely unexplored. Therefore, we propose EchoPrompt, a training-free detector based on latent prompt restoration. Our key intuition is that machine-generated text is typically produced conditioned on an upstream prompt, and this hidden dependency can be partially reactivated by prepending a unified generic prefix. Specifically, EchoPrompt restores a generic assistant-response context, measures the induced likelihood gain with an instruction-tuned model, calibrates it against the corresponding base model, and aggregates the resulting differences into a score that quantifies latent prompt dependency. Extensive experiments show that EchoPrompt achieves state-of-the-art performance among zero-shot detectors while maintaining strong robustness across challenging attack settings.
\end{abstract}

\section{Introduction}

Large language models (LLMs) have made it possible to generate high-quality text at scale, substantially narrowing the observable gap between machine-generated and human-written content. Modern systems can produce fluent and convincing text across diverse domains, making it increasingly difficult to distinguish machine-generated content from human writing. While these capabilities enable many useful applications in content creation, question answering, and writing assistance, they also raise growing concerns in high-stakes scenarios such as misinformation dissemination, spam and fraud generation, educational misuse, authorship ambiguity, and intellectual property protection~\citep{adelani2020generating, ahmed2021detecting}. Prior work~\citep{clark2021all} has further shown that humans themselves often struggle to reliably distinguish model-generated text from human writing. These developments make robust detection of machine-generated text increasingly essential for the safe and trustworthy deployment of LLM systems.

Theoretical analyses~\citep{chakraborty2023possibilities} suggest that AI-generated text detection remains feasible when sufficient textual evidence is available. Existing detectors are commonly divided into training-based and training-free methods. Training-based methods learn discriminative representations from labeled examples, but their dependence on task-specific data can limit scalability and cross-domain generalization~\citep{uchendu2020authorship,li2024mage}. Training-free methods instead derive detection scores from proxy-model statistics, such as token likelihoods, probability curvature, and cross-model discrepancies \citep{mitchell2023detectgpt,hans2024spotting,liu2026zero}. However, these methods typically evaluate the observed text without explicitly probing its dependence on the missing generation context. Instruction-tuned LLMs are trained to produce responses conditioned on prompts, yet these prompts are usually unavailable at detection time. This raises a question: does prompt-conditioned generation leave a persistent signal that remains detectable after the original prompt is removed?

We hypothesize that this persistent signal comes from the latent assistant-response context introduced by instruction tuning. Unlike human-written text, machine-generated text is typically produced as a response under a system prompt, a user request, or an assistant role. Although this original prompt is removed, its influence is not fully erased: the generated text still implicitly ``remembers'' that it was written as a response. Surprisingly, we find that this hidden dependency can be reactivated without knowing the true prompt. In particular, simply prepending a unified generic assistant-style prefix makes machine-generated text align more naturally with the induced conditioning context, whereas human-written text exhibits a much weaker response to the same prefix.

Based on this observation, we propose EchoPrompt, a training-free detector that measures latent prompt dependency through restored assistant-context scoring. EchoPrompt first prepends a unified assistant-style prefix to the input text to restore a generic assistant-response context. It then computes token-level likelihoods using an instruction-tuned proxy model and calibrates the context-induced gain with the corresponding base model, thereby reducing the effect of ordinary linguistic regularities. Finally, these calibrated likelihood differences are aggregated into a sequence-level detection score, where a larger score indicates stronger compatibility with the restored assistant context and thus a higher likelihood of machine generation.

To evaluate the effectiveness of EchoPrompt, we compare it against representative training-based and training-free baselines across three benchmark datasets. Experimental results demonstrate that EchoPrompt consistently achieves stronger overall performance and exhibits favorable robustness across diverse evaluation settings. Further analyses support the contribution of explicitly modeling latent prompt dependency to these improvements, as machine-generated text exhibits greater context-induced likelihood gains even within a single base model (Section \ref{subsec:contextual_restoration_validation}).

\noindent\textbf{Contributions.}
This study makes three main contributions:
\begin{itemize}
    \item We identify latent prompt dependency as a discriminative signal for machine-generated text detection, showing that restoring a generic assistant-response context induces larger likelihood gains for machine-generated text than for human-written text, even without access to the original prompt.
    
    \item We propose EchoPrompt, a training-free detector that restores a generic assistant-response context and measures context-conditioned likelihood gain through calibrated comparison between a base model and an instruction-tuned proxy model.

    \item Extensive experiments across three public benchmarks show that EchoPrompt consistently improves zero-shot detection across proxy models and challenging attack settings. Additional analyses demonstrate strong performance in the low-false-positive-rate regime, transfer of the same fixed prefix across model families, and robustness to continuation-style and human-style generation.
\end{itemize}

\section{Related Work}

\noindent\textbf{Training-based detectors.}
Training-based methods require large-scale labeled examples and supervised models to learn discriminative representations for separating human-written and AI-generated text. Early systems such as OpenAI's GPT-2 output detector~\citep{solaiman2019release} fine-tune pretrained encoders such as RoBERTa to distinguish generated and human text in representation space. Subsequent work improves this paradigm through stronger learning objectives and more structured representation learning. RADAR~\citep{hu2023radar} adopts adversarial learning to improve robustness against paraphrased inputs; BiScope~\citep{guo2024biscope} introduces bidirectional cross-entropy statistics to capture both forward token prediction and preceding-token memorization; DeTeCtive~\citep{guo2024detective} uses multi-level contrastive learning to separate writing styles from different sources; and DETree~\citep{he2025detree} models human--AI collaborative writing processes with tree-structured hierarchical representation learning. Beyond classifier-based designs, R-Detect~\citep{song2025deep} introduces a deep kernel relative test to reduce false positives under distributional mismatch. However, supervised or reference-set-based detectors can still be sensitive to domain, style, generator, and attack shifts, as shown in prior studies on neural text attribution and robustness-oriented detection benchmarks~\citep{uchendu2020authorship,chakraborty-etal-2023-counter}. These limitations motivate more scalable and generalizable training-free alternatives.

\noindent\textbf{Training-free (zero-shot) detectors.}
Training-free methods avoid fitting a task-specific classifier and instead convert token-level statistical signals extracted from the generation distributions of proxy language models into detection scores. Early zero-shot approaches, including LogRank~\citep{gehrmann2019gltr}, Likelihood~\citep{hashimoto2019unifying}, and Entropy~\citep{ippolito2020automatic}, use uncertainty- or rank-based statistics to capture regularities of generated text. DetectLLM~\citep{su2023detectllm} further leverages log-rank information to improve zero-shot detection of machine-generated text. Another line studies the likelihood landscape: DetectGPT~\citep{mitchell2023detectgpt} estimates probability curvature through random perturbations, while Fast-DetectGPT~\citep{bao2023fast} replaces expensive perturbation with conditional probability curvature approximation for substantially improved efficiency. More recent methods exploit cross-model, alignment-aware, or sequence-level signals. Binoculars~\citep{hans2024spotting} compares paired observer and performer models through perplexity-based ratios; IRM~\citep{liu2026zero} derives an implicit reward signal from base and instruction-tuned model pairs without preference collection or additional training; LastDE and LastDE++~\citep{xu2024training} mine token probability sequences for local and global diversity-entropy statistics; and DNA-DetectLLM~\citep{zhu2025dna} models the repair effort required to transform a text toward an ideal machine-generated sequence.

\section{Method}
\label{sec:method}

\subsection{Preliminary}
\label{subsec:preliminary}

\noindent\textbf{Pre-training workflow of LLM.}
Modern LLMs typically undergo a multi-stage optimization pipeline. LLMs are first pre-trained as autoregressive language models over open-domain corpora. Given a token sequence $X=(x_1,\dots,x_n)$, a base model factorizes its likelihood as
\begin{equation}
    P_{\mathrm{base}}(X)
    = \prod_{t=1}^{n} P_{\mathrm{base}}(x_t \mid x_{<t}).
    \label{eq:base_likelihood}
\end{equation}
The corresponding pre-training objective minimizes the negative log-likelihood:
\begin{equation}
    \theta_{\mathrm{base}}^{*}
    = \arg\min_{\theta}
    \left[
    -\mathbb{E}_{X\sim\mathcal{D}_{\mathrm{pre}}}
    \sum_{t=1}^{n}\log P_{\theta}(x_t\mid x_{<t})
    \right].
    \label{eq:base_objective}
\end{equation}
This stage mainly captures general linguistic regularities, without explicitly optimizing the model to generate responses under user instructions~\citep{brown2020language}.

\noindent\textbf{Post-training workflow of LLM.}
    
Subsequently, the model undergoes post-training, typically including supervised fine-tuning (SFT) and preference-based optimization such as RLHF~\citep{ouyang2022training} or DPO~\citep{rafailov2023direct}. Here, the optimization objective shifts from unsupervised continuation to conditional generation under a global instruction $c_g$:
\begin{equation}
    P_{\text{inst}}(X \mid c_g)
    = \prod_{t=1}^{n} P_{\text{inst}}(x_t \mid c_g, x_{<t}).
\end{equation}
The shared effect of SFT and preference-based post-training can be schematically summarized by the following unified objective:

\begin{equation}
\begin{aligned}
\theta_{\text{inst}}^{*}
= \arg\min_{\theta}\Big[
&\mathcal{L}_{\text{SFT}}(\theta)
+ \alpha\,\mathcal{L}_{\text{pref}}(\theta)+ \lambda\,\mathrm{KL}\!\left(\pi_{\theta}\,\|\,\pi_{\text{base}}\right)
\Big],
\end{aligned}
\end{equation}
where
{\small
\begin{equation}
\mathcal{L}_{\text{SFT}}(\theta)=-\mathbb{E}_{(c_g,X)\sim\mathcal{D}_{\text{SFT}}}\left[\sum_{t=1}^{n}\log P_{\theta}(x_t\mid c_g,x_{<t})\right],
\end{equation}
}
and $\mathcal{L}_{\text{pref}}$ denotes a preference-based objective, and the above formulation provides a schematic abstraction of common post-training procedures. From the perspective of the Transformer architecture~\citep{vaswani2017attention}, once the instruction prefix $c_g$ is prepended to the sequence, it becomes part of the causal context. Its token representations can be attended to by subsequent tokens and propagated through the network, thereby influencing downstream hidden states and decoding decisions.

\subsection{EchoPrompt}
\label{subsec:echoprompt}

\begin{figure}[t]
  \centering
  \includegraphics[width=0.98\linewidth]{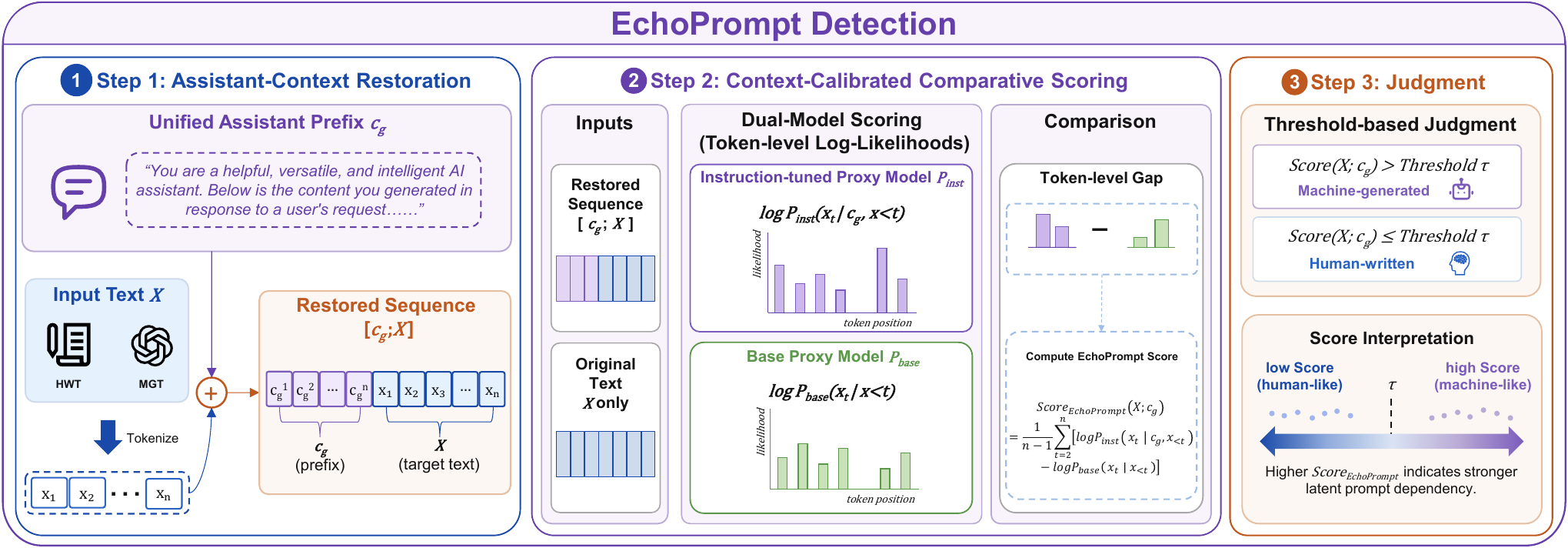}
  \caption{
  Overview of EchoPrompt. 
  }
  \label{fig:echoprompt_overview}
\end{figure}

EchoPrompt is a training-free detector that probes whether a target passage exhibits an unusually strong dependency on a restored assistant-style context. Figure~\ref{fig:echoprompt_overview} provides an overview of EchoPrompt. The detection process consists of three steps:

\noindent\textbf{Step 1: Assistant-context restoration.}
Given an input text $X$, EchoPrompt prepends a unified task-agnostic assistant-style prefix $c_g$ to construct a restored sequence $[c_g;X]$, which approximates the generic response condition under which AI-generated text is commonly produced.

\noindent\textbf{Step 2: Context-calibrated comparative scoring.}
EchoPrompt computes token-level log-likelihoods under two asymmetric conditions: the instruction-tuned proxy model evaluates the restored sequence $[c_g;X]$, while the corresponding base model evaluates the original text $X$ to calibrate ordinary linguistic predictability. Their difference defines a context-calibrated token-level gap, and the average gap forms the sequence-level EchoPrompt score.

\noindent\textbf{Step 3: Threshold-based detection.}
The final prediction is made by comparing the EchoPrompt score with a threshold $\tau$: passages with scores above the threshold are classified as AI-generated, while the remaining passages are classified as human-written.

\subsubsection{Assistant-Context Restoration}
\label{subsec:contextual_restoration}

EchoPrompt is motivated by a simple hypothesis: machine-generated text is typically produced under an implicit assistant-style directive, even when the original prompt is unavailable at detection time. Therefore, if we restore a generic assistant-response context before evaluating the text, machine-generated passages should align with that condition more naturally than human-written passages.

To operationalize this idea, we prepend a global prefix $c_g$ to each evaluated passage to restore a generic assistant-response context. %
The choice of $c_g$ is determined through preliminary evaluation on an independent development set from HC3~\citep{guo2023close}, which is disjoint from all evaluation benchmarks. Once selected, the same prefix is fixed across all datasets and proxy model families. Detailed settings and results are provided in Appendix~\ref{app:prefix-design}. 
Based on these results, we instantiate $c_g$ as:
\begin{quote}
    \small \textit{``You are a helpful, versatile, and intelligent AI assistant. Below is the content you generated in response to a user's request, which acts as either a coherent continuation, a topic-specific article, or a detailed answer to a question:\textbackslash n\textbackslash n''}
\end{quote}
This prefix is intentionally task-agnostic. Rather than introducing specific entities or task instructions, it restores only the coarse global condition that the following sequence should be interpreted as an assistant-style response. In this way, the detector does not rely on access to the original prompt, but instead probes whether the target text is inherently compatible with a generic assistant-response context.

\subsubsection{Context-Calibrated Comparative Scoring}
\label{subsec:context_calibrated_scoring}

After restoring the assistant-style context, a natural option is to directly measure the conditional likelihood under the instruction-tuned model, i.e., $\log P_{\text{inst}}(x_t \mid c_g, x_{<t})$. However, this quantity alone is not sufficiently discriminative: high-frequency tokens, common phrases, and the intrinsic fluency of the text can all increase token probabilities, making it difficult to separate true context dependency from ordinary local smoothness.

To reduce this confounding effect, EchoPrompt introduces a calibrated comparison against a base model. The base model primarily captures the marginal regularities of open-domain text, and thus serves as a reference for local linguistic predictability without the restored assistant-style conditioning. Based on this contrast, we define the EchoPrompt score as:
\begin{equation}
\begin{aligned}
Score_{\text{EchoPrompt}}(X; c_g)= \frac{1}{n-1} \sum_{t=2}^{n}
\Big[
\log P_{\text{inst}}(x_t \mid c_g, x_{<t}) - \log P_{\text{base}}(x_t \mid x_{<t})
\Big],
\end{aligned}
\end{equation}
where $X$ denotes the evaluated sequence and $c_g$ denotes the restored assistant-style prefix. The first term measures how naturally the passage is supported under assistant-style contextual conditioning, while the second term provides a calibration baseline for its local linguistic predictability. Their difference suppresses fluency effects shared by both human and machine text, and highlights the additional advantage that machine-generated passages receive when evaluated under the restored assistant-style condition. Averaging across token positions yields a stable sequence-level statistic for zero-shot detection.

\subsubsection{Threshold-based Detection}
\label{subsec:threshold_detection}
EchoPrompt classifies the evaluated sequence by comparing its score with a threshold $\tau$:
\begin{equation}
\mathcal{D}(X)=
\begin{cases}
\text{AI-generated Text}, & Score_{\text{EchoPrompt}}(X;c_g) > \tau,\\
\text{Human-written Text}, & Score_{\text{EchoPrompt}}(X;c_g) \leq \tau.
\end{cases}
\end{equation}
A higher EchoPrompt score indicates stronger latent dependency on the restored assistant-style context, and therefore a higher likelihood of machine generation.

\subsection{Empirical Evidence of Latent Prompt Dependency}
\label{subsec:contextual_restoration_validation}

To empirically validate the role of the restored prefix itself, we directly examine the likelihood gain induced by prompt injection within the \textit{same model}. For a fixed model $P$, we define
\begin{equation}
\begin{aligned}
g(X; c_g)= \frac{1}{n-1}\sum_{t=2}^{n}
\Big[
\log P(x_t\mid c_g, x_{<t}) - \log P(x_t\mid x_{<t})
\Big],
\end{aligned}
\end{equation}
which measures how much the target sequence benefits from the restored assistant-style context.

\begin{figure*}[h]
\centering
\includegraphics[width= 0.98\textwidth]{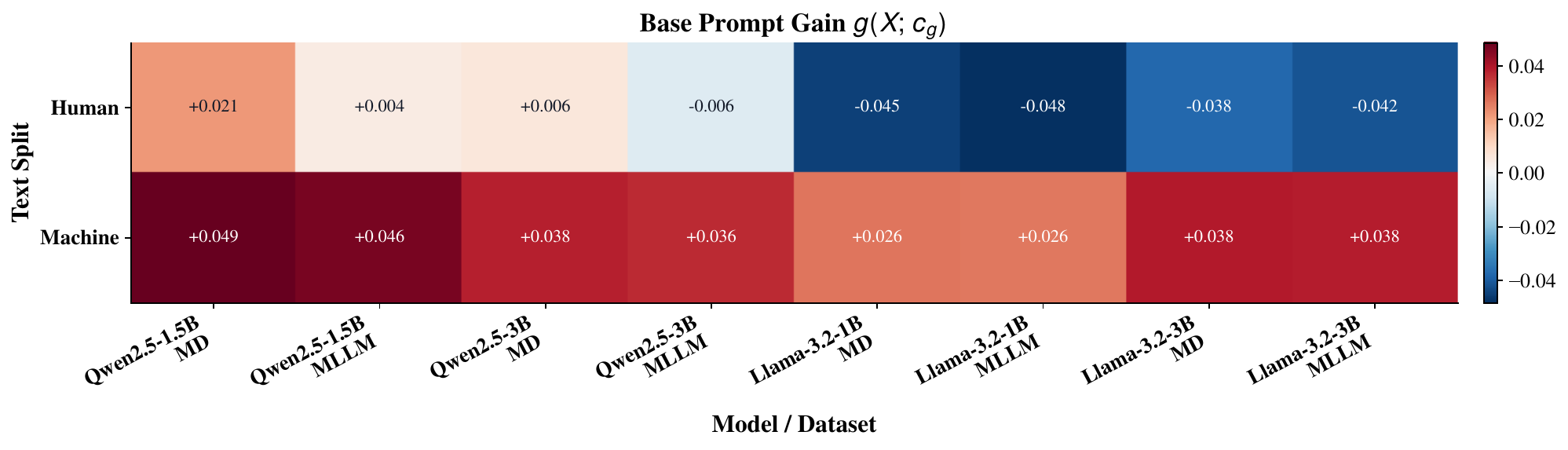}
\caption{Prompt-induced contextual gain under a restored assistant-style prefix for the base models of four small proxy families.}
\label{fig:contextual-restoration-gain}
\end{figure*}

Figure~\ref{fig:contextual-restoration-gain} reports the average gain for human and machine text under the base models of four small proxy families, and Appendix~\ref{app:Context-Conditioned Gain} provides the detailed settings and values. A consistent pattern emerges: after the same generic prefix is injected, machine-generated text receives a larger likelihood gain than human-written text.
Since this comparison is performed within the same base model, it directly isolates the effect of context restoration. The consistent gain difference provides empirical evidence that machine-generated text responds differently to the restored assistant-response context, supporting the motivation for explicitly modeling latent prompt dependency in EchoPrompt.

\section{Experiments}
\label{sec:experiments}

\subsection{Experimental Setup}
\label{sec:exp-setup}

We evaluate EchoPrompt on three public benchmarks, DetectRL~\citep{wu2024detectrl}, RealDet~\citep{zhu2025reliably}, and RAID~\citep{dugan2024raid}, against representative training-based and training-free detectors. Unless otherwise specified, the main comparisons use the Llama-3-8B base/instruction pair as the proxy models, while additional proxy families and scales are evaluated in our cross-proxy analysis. We report AUROC and Best-F1 as the primary metrics, together with TPR@1\%FPR for low-false-positive-rate evaluation. For fair comparison, all training-based detectors are trained on HC3~\citep{guo2023close}, which is disjoint from the evaluation benchmarks. Full dataset statistics, baseline descriptions, proxy-model settings, and implementation details are provided in Appendix~\ref{app:experimental-setup}.

\begin{table*}[t]
\centering
\caption{Performance comparison (\%) across different benchmarks. Bold and underlines mark the best and second-best results within each training-free proxy-model block.}
\label{tab:main-results-llama}
\scriptsize
\setlength{\tabcolsep}{2.8pt}
\renewcommand{\arraystretch}{1.03}
\resizebox{\textwidth}{!}{%
\begin{tabular}{lcccccccccccc}
\toprule
Detectors & \multicolumn{2}{c}{\shortstack{\textbf{DetectRL}\\\textbf{Multi-Domain}}} & \multicolumn{2}{c}{\shortstack{\textbf{DetectRL}\\\textbf{Multi-LLM}}} & \multicolumn{2}{c}{\shortstack{\textbf{DetectRL}\\\textbf{Multi-Attack}}} & \multicolumn{2}{c}{\textbf{RealDet}} & \multicolumn{2}{c}{\textbf{RAID}} & \multicolumn{2}{c}{\textbf{Avg}} \\
\cmidrule(lr){2-3}\cmidrule(lr){4-5}\cmidrule(lr){6-7}\cmidrule(lr){8-9}\cmidrule(lr){10-11}\cmidrule(lr){12-13}
& AUROC & Best-F1 & AUROC & Best-F1 & AUROC & Best-F1 & AUROC & Best-F1 & AUROC & Best-F1 & AUROC & Best-F1 \\
\midrule
\methodgrouprow{13}{Training-based Methods}
OpenAI-D & 82.64 & 76.00 & 82.58 & 75.61 & 83.91 & 78.02 & 90.06 & 77.15 & 73.03 & 67.19 & 82.44 & 74.80 \\
BiScope & 76.81 & 70.75 & 76.91 & 70.37 & 79.10 & 74.07 & 88.90 & 72.08 & 79.42 & 73.93 & 80.23 & 72.24 \\
R-Detect & 77.83 & 73.29 & 78.22 & 73.52 & 78.54 & 74.18 & 87.84 & 72.73 & 70.75 & 68.39 & 78.64 & 72.42 \\
\midrule
\methodgrouprow{13}{Training-free Methods}
\textit{Llama-3-8B family} & \multicolumn{12}{l}{} \\
Entropy & 64.31 & 66.71 & 64.14 & 66.68 & 67.64 & 71.06 & 77.08 & 73.04 & 67.73 & 66.69 & 68.18 & 68.84 \\
Likelihood & 79.54 & 73.57 & 79.24 & 72.75 & 81.71 & 76.12 & 85.53 & 81.18 & 75.89 & 73.11 & 80.38 & 75.34 \\
LogRank & 76.83 & 70.70 & 76.39 & 69.63 & 79.17 & 74.63 & 85.34 & 80.81 & 75.80 & 72.49 & 78.71 & 73.65 \\
Fast-DetectGPT & 91.40 & 84.02 & 91.45 & 83.61 & 92.63 & 85.83 & 88.90 & 82.96 & 83.68 & 83.01 & 89.61 & 83.89 \\
Binoculars & 91.93 & 84.98 & 92.01 & 84.57 & 93.10 & 86.61 & 89.58 & \underline{84.46} & 83.73 & 82.49 & 90.07 & 84.62 \\
LastDE++ & 84.26 & 76.72 & 84.26 & 76.74 & 86.69 & 80.43 & 89.10 & 81.74 & \underline{87.81} & \underline{83.94} & 86.42 & 79.91 \\
DNA-DetectLLM & 90.41 & 83.28 & 90.55 & 82.99 & 91.86 & 85.05 & 87.75 & 82.42 & 80.94 & 80.55 & 88.30 & 82.86 \\
IRM & \underline{98.47} & \underline{94.06} & \underline{98.50} & \underline{93.89} & \underline{98.36} & \underline{93.92} & \underline{91.51} & 83.43 & 87.52 & 81.42 & \underline{94.87} & \underline{89.34} \\
\textbf{EchoPrompt} & \textbf{98.82} & \textbf{95.26} & \textbf{98.62} & \textbf{95.25} & \textbf{98.60} & \textbf{95.28} & \textbf{92.33} & \textbf{88.79} & \textbf{89.45} & \textbf{85.32} & \textbf{95.56} & \textbf{91.98} \\
\bottomrule
\end{tabular}
}
\end{table*}

\subsection{Main Results}

Table~\ref{tab:main-results-llama} shows that EchoPrompt achieves the strongest overall performance under the Llama-3-8B proxy family. The full results are provided in Appendix~\ref{app:Full Main Results}. Across the five evaluation settings, EchoPrompt ranks first on both AUROC and Best-F1, improving over the strongest training-free baseline, IRM, by \textbf{0.69\%} AUROC and \textbf{2.64\%} Best-F1 on average. Although the AUROC margin is relatively small under this strong proxy, the advantage becomes substantially larger when the false-positive rate is strictly controlled. EchoPrompt achieves a macro-average TPR@1\%FPR of \textbf{81.86\%}, compared with \textbf{73.11\%} for IRM, which is particularly important in practical settings where false positives on human-written text need to be kept low. Detailed low-FPR results are provided in Appendix~\ref{app:low-fpr}. Compared with the best training-based detector, OpenAI-D, the gains are much larger, reaching \textbf{13.12\%} AUROC and \textbf{17.18\%} Best-F1 on average. The advantage is also clear on the more distributionally different RealDet and RAID benchmarks: EchoPrompt improves over the second-best method by \textbf{4.33\%} Best-F1 on RealDet and by \textbf{1.64\%} AUROC / \textbf{1.38\%} Best-F1 on RAID. These results indicate that EchoPrompt provides a more effective and transferable detection signal than the strongest existing training-free competitors.

This advantage stems from the motivation of EchoPrompt. Machine-generated text is usually written as an assistant-style response, but conventional zero-shot detectors evaluate it without this missing context, so their signals are easily mixed with generic fluency and token-level regularities. By restoring a task-agnostic assistant-response context, EchoPrompt exposes this latent generation dependency. The base-model comparison further filters out ordinary linguistic predictability, leaving a cleaner signal of assistant-style contextual compatibility. This explains why EchoPrompt generalizes better than strong training-free baselines: it detects not only whether a passage is statistically fluent, but whether it behaves like text generated under an implicit assistant-response condition.

\subsection{Robustness Against Various Attacks}
\begin{table*}[t]
\centering
\caption{Per-attack results (\%) including EchoPrompt and other baselines using the Llama-3-8B family as proxy models across different attack types. Bold and underlines mark the best and second-best results within each training-free proxy-model block.}
\label{tab:attack-results-llama}
\resizebox{\textwidth}{!}{%
\begin{tabular}{lcccccccccc}
\toprule
Detectors & \multicolumn{2}{c}{\textbf{Direct Prompt}} & \multicolumn{2}{c}{\textbf{Prompt Attacks}} & \multicolumn{2}{c}{\textbf{Paraphrase}} & \multicolumn{2}{c}{\textbf{Perturbation}} & \multicolumn{2}{c}{\textbf{Data Mixing}} \\
\cmidrule(lr){2-3}\cmidrule(lr){4-5}\cmidrule(lr){6-7}\cmidrule(lr){8-9}\cmidrule(lr){10-11}
& AUROC & Best-F1 & AUROC & Best-F1 & AUROC & Best-F1 & AUROC & Best-F1 & AUROC & Best-F1 \\
\midrule
\methodgrouprow{11}{Training-based Methods}
OpenAI-D & 93.12 & 83.65 & 88.08 & 78.71 & 87.48 & 79.12 & 71.00 & 72.01 & 79.85 & 76.62 \\
BiScope & 93.61 & 86.43 & 91.02 & 83.06 & 71.68 & 66.67 & 67.96 & 67.54 & 71.24 & 66.67 \\
R-Detect & 90.09 & 80.34 & 85.31 & 76.96 & 86.76 & 78.16 & 62.09 & 68.76 & 68.46 & 66.69 \\
\midrule
\methodgrouprow{11}{Training-free Methods}
\textit{Llama-3-8B family} & \multicolumn{10}{l}{} \\
Entropy & 86.37 & 78.37 & 84.39 & 76.73 & 56.00 & 66.67 & 52.02 & 66.87 & 59.43 & 66.67 \\
Likelihood & 95.64 & 89.39 & 92.33 & 85.35 & 73.31 & 68.66 & 71.77 & 67.60 & 75.48 & 69.59 \\
LogRank & 94.86 & 88.26 & 91.09 & 83.94 & 70.75 & 66.67 & 66.79 & 67.20 & 72.36 & 67.08 \\
Fast-DetectGPT & 97.45 & 92.52 & 95.32 & 89.43 & 90.84 & 83.68 & 88.22 & 80.16 & 91.31 & 83.36 \\
Binoculars & 97.89 & 93.30 & 95.91 & 90.88 & 91.15 & 84.11 & 88.81 & 80.78 & 91.72 & 83.96 \\
LastDE++ & 96.72 & 90.50 & 93.91 & 87.95 & 88.59 & 81.03 & 70.35 & 66.75 & 83.87 & 75.92 \\
DNA-DetectLLM & 97.40 & 92.35 & 95.84 & 90.40 & 87.24 & 79.60 & 88.36 & 79.96 & 90.48 & 82.93 \\
IRM & \underline{98.70} & \underline{94.59} & \underline{97.73} & \underline{92.76} & \textbf{99.15} & \textbf{95.63} & \underline{98.90} & \underline{94.84} & \underline{97.31} & \underline{91.76} \\
\textbf{EchoPrompt} & \textbf{99.62} & \textbf{97.33} & \textbf{98.13} & \textbf{94.81} & \underline{98.28} & \underline{94.53} & \textbf{99.24} & \textbf{97.04} & \textbf{97.74} & \textbf{92.70} \\
\bottomrule
\end{tabular}
}
\end{table*}

Table~\ref{tab:attack-results-llama} further shows that EchoPrompt remains robust across different attack settings. The complete attack results are provided in Appendix~\ref{app:Full Per-Attack Results}. It obtains the best AUROC and Best-F1 in four out of five attack groups and achieves the strongest average attack performance, improving over IRM by \textbf{0.24\%} AUROC and \textbf{1.37\%} Best-F1 on average. The gains are particularly clear under direct prompting and perturbation, where EchoPrompt improves Best-F1 over the second-best method by \textbf{2.74\%} and \textbf{2.20\%}, respectively. It also remains strongest under prompt attacks and data mixing, with Best-F1 gains of \textbf{2.05\%} and \textbf{0.94\%}. Under paraphrasing, EchoPrompt ranks second but remains very close to IRM, trailing by only \textbf{0.87\%} AUROC and \textbf{1.10\%} Best-F1 while still achieving a high Best-F1 of \textbf{94.53\%}. This robustness pattern is consistent with the design of EchoPrompt. Prompt attacks can alter superficial prompting cues, data mixing can dilute local token statistics, and perturbation can directly corrupt lexical-level regularities. Methods that rely mainly on unconditional likelihood, entropy, rank, or local model discrepancy are therefore more easily affected by these transformations. In contrast, EchoPrompt focuses on whether the text retains generation-style dependency rather than on isolated token statistics. This makes the context-calibrated signal both discriminative and stable under diverse adversarial transformations.

\noindent\textbf{Robustness Beyond Standard Attacks.}
We further evaluate EchoPrompt under stronger attacks that alter the generation task, explicitly encourage human-like writing, or rewrite the generated text.
On DetectRL~\citep{wu2024detectrl}, Yelp \texttt{direct\_prompt} changes the task into continuation writing from a human-written opening sentence, while Few-shot and ICO/SICO explicitly encourage the model to imitate human writing. EchoPrompt still achieves AUROCs of \textbf{99.98\%}, \textbf{96.96\%}, and \textbf{99.60\%} under these three settings. It also reaches \textbf{98.46\%} AUROC after back-translation rewriting in DetectRL~\citep{wu2024detectrl}. These transformations substantially change the surface form of the generated text, but they do not remove the post-training instruction-following process that produced it, and the detection signal therefore remains strong.

We also evaluate non-chat/completion outputs from RAID~\citep{dugan2024raid}, including GPT-2, GPT-3, Cohere, MPT, and Mistral, where EchoPrompt obtains an AUROC of \textbf{82.50\%}. The expected performance drop in this setting is consistent with our hypothesis: generation that does not rely on instruction-tuned assistant behavior may contain a weaker latent dependency on the response context exploited by EchoPrompt. Detailed results are provided in Appendix~\ref{app:targeted-robustness}.

\subsection{Ablation Study}
\label{subsec:ablation}
\noindent\textbf{Effect of Prompt Choice.} 

We conduct a prompt-component ablation to study the effect of the generic prefix on detection performance. Detailed settings are provided in Appendix~\ref{app:prompt-ablation}. As shown in Figure~\ref{fig:prompt-ablation}, the full prefix consistently outperforms the empty-prompt setting, improving AUROC by \textbf{14.73\%} and \textbf{12.57\%} on Qwen2.5-3B, and by \textbf{5.63\%} and \textbf{5.33\%} on Llama-3.1-8B. These gains verify the key motivation of EchoPrompt: even without access to the original user prompt, machine-generated text is more naturally compatible with a restored assistant-response condition than human-written text.

The component-level results further show that the gain does not simply come from assigning an assistant identity to the proxy model. The role sentence alone brings only limited gains, especially on Llama-3.1-8B, where the improvements are about \textbf{1.60\%} and \textbf{1.44\%}. In contrast, context clause A, which frames the passage as content generated in response to a user's request, provides the strongest individual contribution, improving AUROC by \textbf{12.63\%}, \textbf{9.28\%}, \textbf{3.84\%}, and \textbf{3.41\%} across the four settings. The leave-one-out results lead to the same conclusion. This indicates that the core signal of EchoPrompt comes from restoring the missing prompt--response relation, while the role sentence and context clause B mainly help stabilize and broaden this generic assistant-style condition.

\begin{figure*}[t]
\centering
\includegraphics[width=0.96\textwidth]{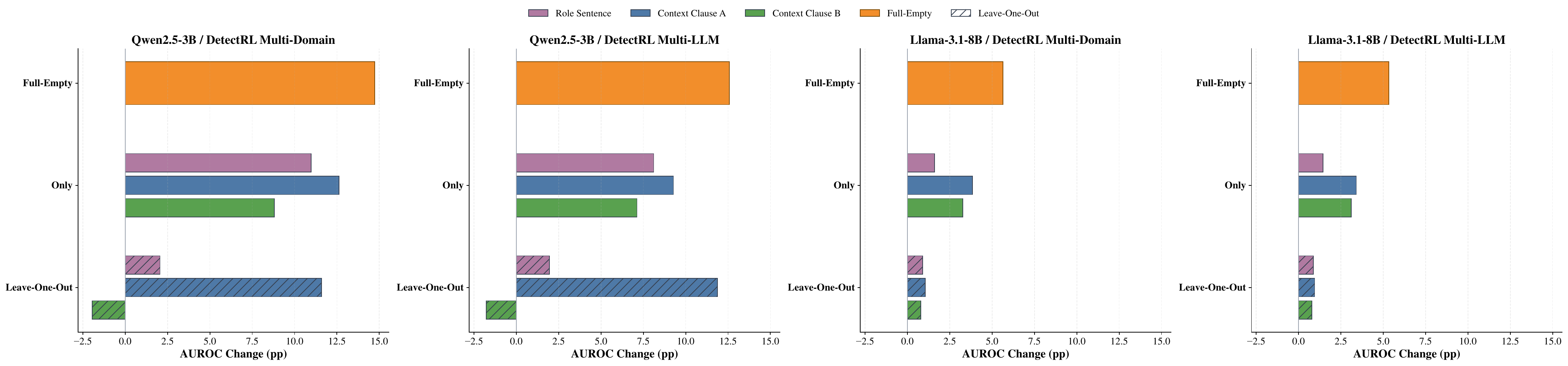}
\caption{Prompt-component ablation results.}
\label{fig:prompt-ablation}
\end{figure*}

\begin{figure*}[t]
\centering

\begin{minipage}[h]{0.52\textwidth}
    \centering
    \includegraphics[width=\linewidth]{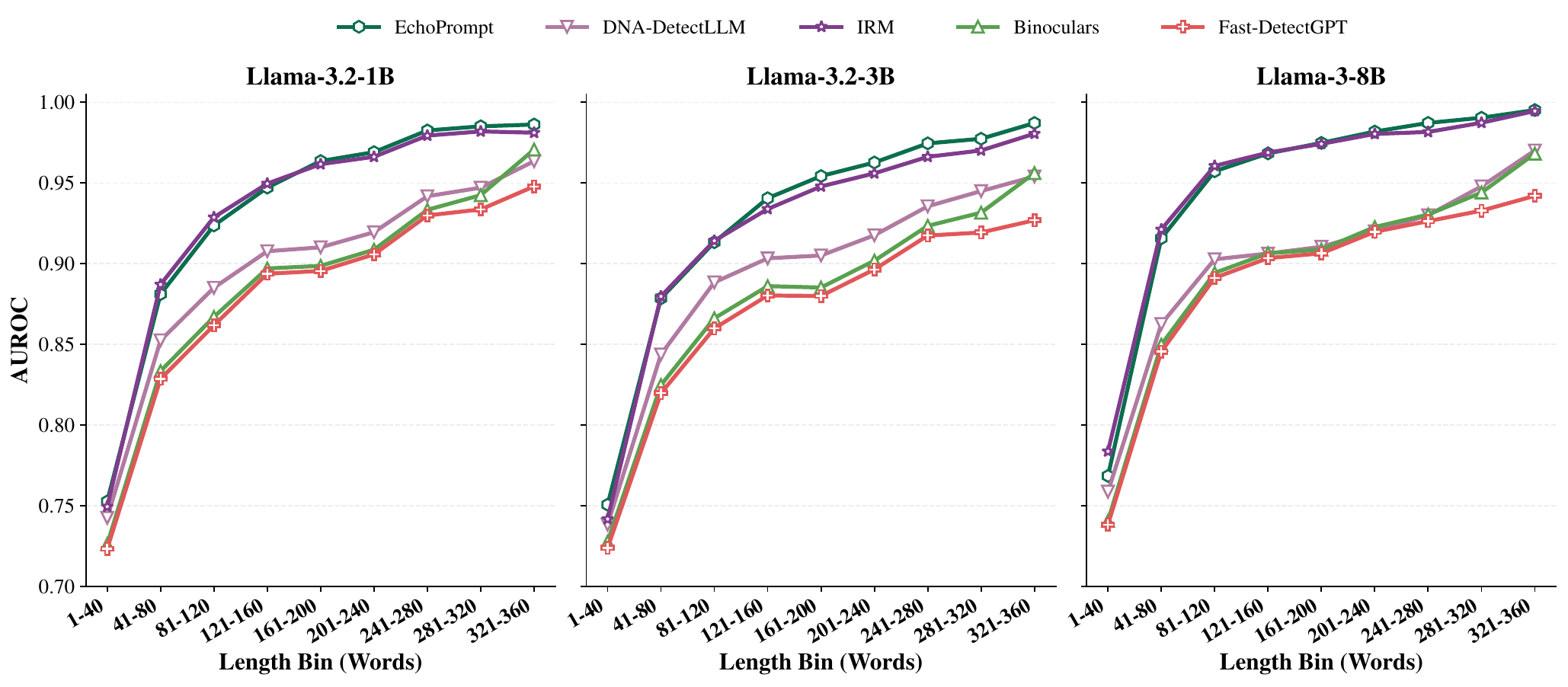}
    \caption{Length-binned AUROC variation of different zero-shot detectors on DetectRL Length.}
    \label{fig:length-auroc-llama}
\end{minipage}
\hfill
\begin{minipage}[h]{0.46\textwidth}
    \centering
    \includegraphics[width=\linewidth]{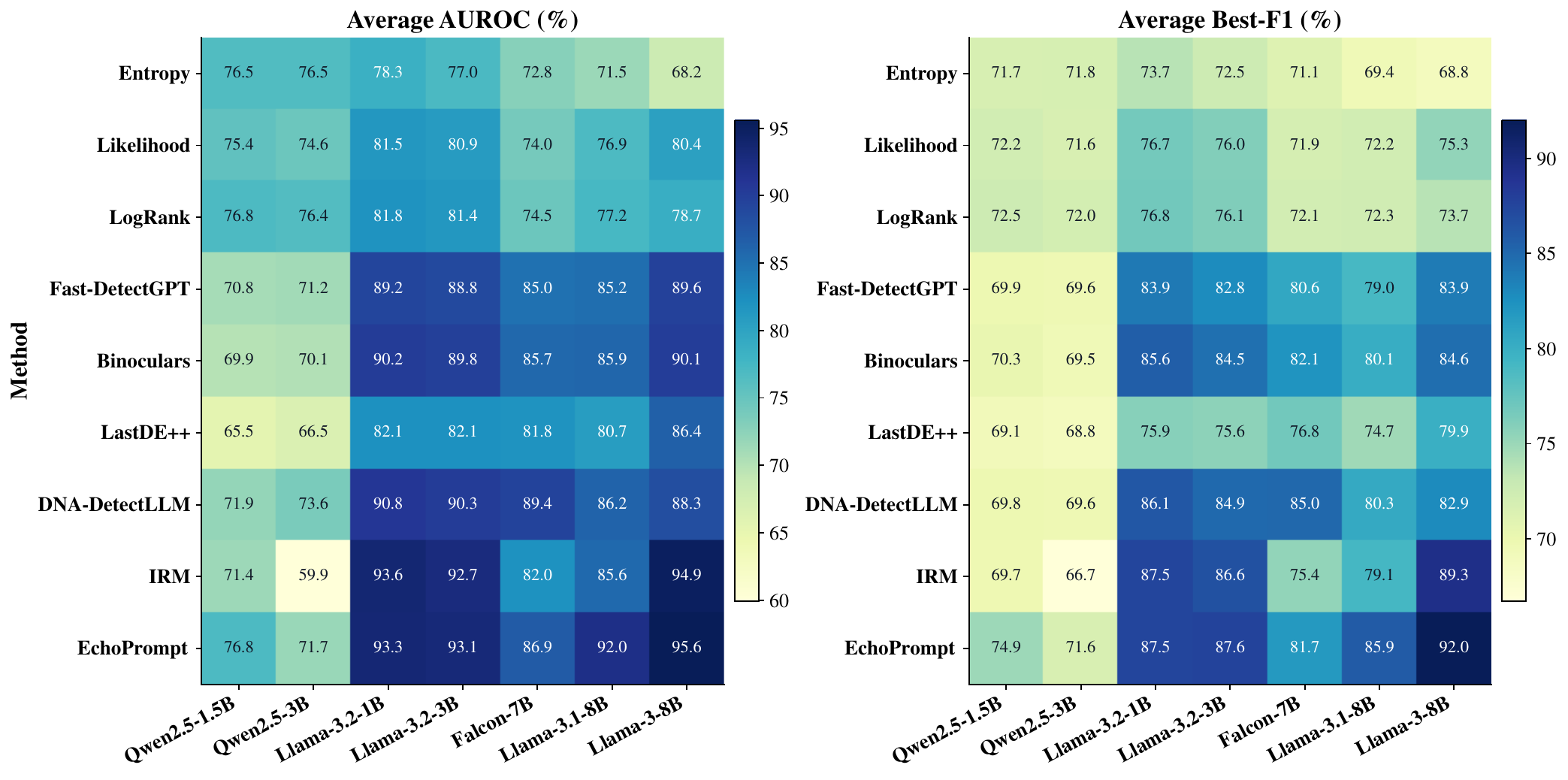}
    \caption{Proxy-model analysis with representative detector families.}
    \label{fig:proxy-model-analysis}
\end{minipage}

\end{figure*}

\noindent\textbf{Effect of Proxy Models.}

Figure~\ref{fig:proxy-model-analysis} shows that EchoPrompt remains strong across different proxy model families and scales. Averaged over all proxy settings, EchoPrompt achieves \textbf{87.06\%} AUROC and \textbf{83.03\%} Best-F1, outperforming the strongest competing average baseline, DNA-DetectLLM, by \textbf{2.70\%} AUROC and \textbf{3.23\%} Best-F1. The advantage is more evident on Llama-family proxies, where EchoPrompt reaches \textbf{93.50\%} AUROC and \textbf{88.25\%} Best-F1 on average, exceeding IRM by \textbf{1.80\%} AUROC and \textbf{2.63\%} Best-F1. These results indicate that EchoPrompt achieves strong cross-proxy robustness, while its performance is still influenced by the specific proxy family selected.

\noindent\textbf{Effect of Text Lengths.}

Figure~\ref{fig:length-auroc-llama} analyzes AUROC across different length bins on DetectRL Length. Short texts are challenging for all detectors because they provide fewer tokens for estimating reliable detection signals. Even in the shortest 1--40 word bin, EchoPrompt remains competitive, with an average AUROC of \textbf{75.8\%} across the three Llama-family proxies. As length increases, its performance rises rapidly to \textbf{93.1\%} in the 81--120 bin and \textbf{98.9\%} in the 321--360 bin. These results highlight that EchoPrompt is able to capture stable discriminative signals across texts of different lengths, thereby achieving strong detection robustness under varying length conditions.

\noindent\textbf{Effect of Scoring Formulation.}
We validate the scoring design of EchoPrompt. To examine the necessity of the asymmetric formulation, we compare the EchoPrompt score, $P_{\mathrm{inst}}(X \mid c_g)-P_{\mathrm{base}}(X)$, with two alternative scoring formulations using the Llama-3-8B base/instruction pair on DetectRL Multi-Domain~\citep{wu2024detectrl}. As shown in Table~\ref{tab:scoring-ablation}, the EchoPrompt formulation achieves the best performance, demonstrating the effectiveness of the proposed scoring design.
\begin{wraptable}{r}{0.55\textwidth}
\vspace{-0.8\baselineskip}
\centering
\caption{Comparison of alternative scorings.}
\label{tab:scoring-ablation}
\small
\setlength{\tabcolsep}{4pt}
\renewcommand{\arraystretch}{1.08}

\begin{tabular}{p{0.53\linewidth}cc}
\toprule
\textbf{Score} & \textbf{AUROC} & \textbf{Best-F1} \\
\midrule

$P_{\mathrm{inst}}(X \mid c_g)-P_{\mathrm{base}}(X)$
& \textbf{0.9864} & \textbf{0.9519} \\

$P_{\mathrm{inst}}(X \mid c_g)
-P_{\mathrm{base}}(X \mid c_g)$
& 0.9683 & 0.9162 \\

$\begin{aligned}
&[P_{\mathrm{base}}(X \mid c_g)-P_{\mathrm{base}}(X)]\\
&\quad-[P_{\mathrm{inst}}(X \mid c_g)-P_{\mathrm{inst}}(X)]
\end{aligned}$
& 0.9194 & 0.8416 \\

\bottomrule
\end{tabular}

\vspace{-0.5\baselineskip}
\end{wraptable}

\noindent\textbf{Effect of Answer Style.}We further evaluate EchoPrompt on human-written but assistant-like text using HC3~\citep{guo2023close}, which contains paired human and ChatGPT answers to the same questions. We sample 1,000 pairs from each of WikiQA, Medical, and FiQA and evaluate them using the Meta-Llama-3-8B base/instruction pair. EchoPrompt achieves AUROC/Best-F1 scores of \textbf{98.02\%}/\textbf{95.76\%} on WikiQA, \textbf{99.87\%}/\textbf{99.15\%} on Medical, and \textbf{98.09\%}/\textbf{95.04\%} on FiQA, with macro averages of \textbf{98.66\%}/\textbf{96.65\%}. These results show that EchoPrompt clearly separates answer-like human text from paired ChatGPT outputs rather than simply classifying answer-style prose as machine-generated, supporting its ability to capture potential instruction dependency: ChatGPT answers are shaped by instruction-conditioned generation, whereas human answers are not.

\subsection{Further Analysis}

\begin{figure*}[t]
\centering

\begin{minipage}[h]{0.62\textwidth}
    \vspace{0pt}
    \centering
    \includegraphics[width=\linewidth]{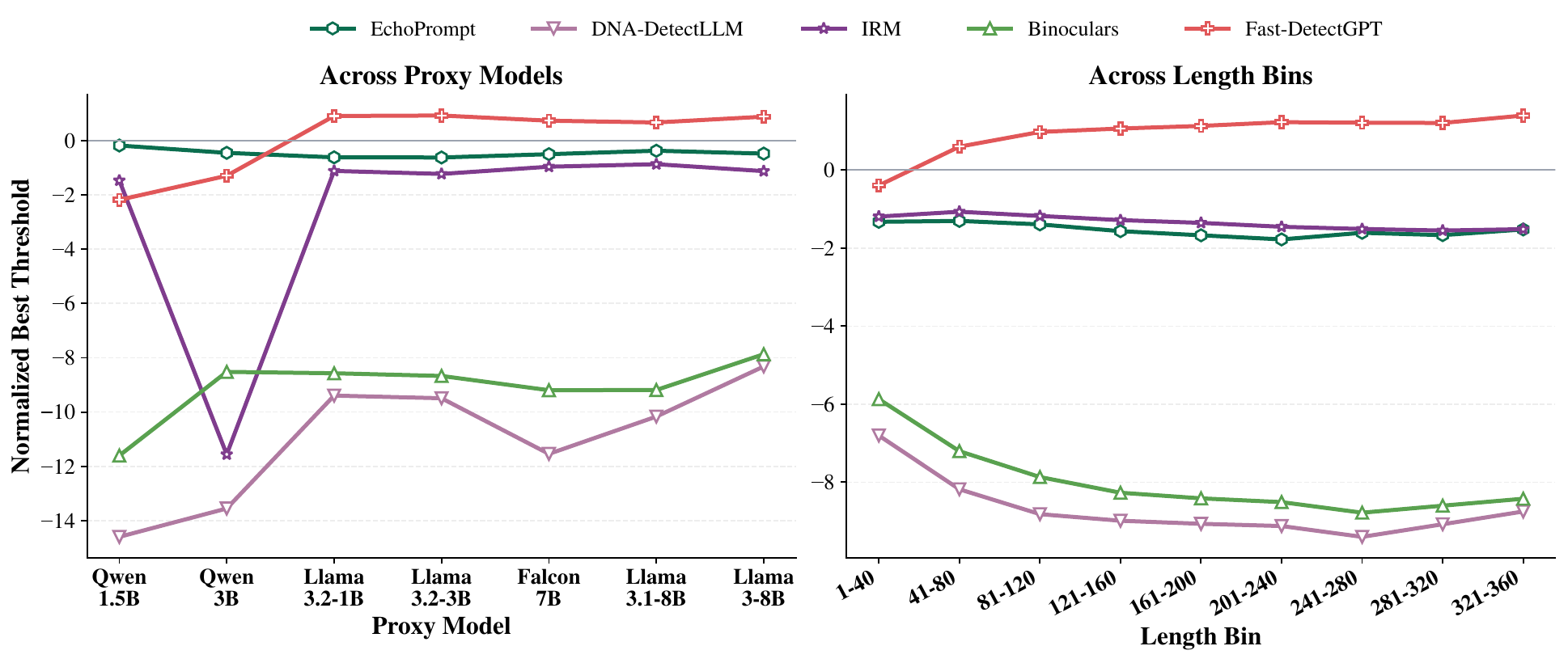}
    \captionof{figure}{Normalized threshold trajectories across proxy models and text-length bins.}
    \label{fig:threshold-stability}
\end{minipage}
\hfill
\begin{minipage}[h]{0.36\textwidth}
    \vspace{0pt}
    \centering
    \includegraphics[width=\linewidth]{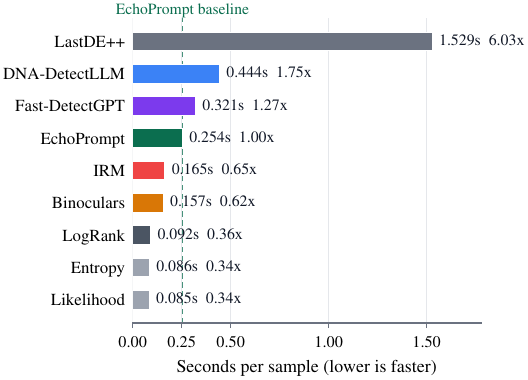}
    \captionof{figure}{Average runtime per sample.}
    \label{fig:efficiency}
\end{minipage}

\end{figure*}

\noindent\textbf{Threshold Stability.}
In practical deployment, a stable decision threshold is important across different proxy models and text lengths. We therefore analyze the normalized threshold $\hat{\tau}=\tau^{*}/\sigma_{\mathrm{pool}}$, where $\tau^{*}$ is the F1-optimal threshold and $\sigma_{\mathrm{pool}}$ is the pooled within-class score standard deviation. The sign of $\hat{\tau}$ is method-dependent, so we focus on whether the trajectories remain smooth and compact after normalization. As shown in Figure~\ref{fig:threshold-stability}, EchoPrompt exhibits one of the most stable profiles across both proxy and length regimes, indicating better threshold stability and easier deployment under changing evaluation conditions.

\noindent\textbf{Efficiency Analysis.}
Figure~\ref{fig:efficiency} reports the average runtime per sample for each detector. Likelihood, Entropy, and LogRank are the fastest methods, while LastDE++ and DNA-DetectLLM are substantially more expensive. EchoPrompt falls in the middle range: although it requires additional computation for context-conditioned scoring, it remains efficient relative to heavier baselines and provides a favorable efficiency--accuracy trade-off.

\section{Conclusion}
In this paper, we proposed EchoPrompt, a training-free detector that identifies machine-generated text by measuring its dependency on restored assistant-style context. By combining generic prefix restoration with calibrated likelihood comparison between base and instruction-tuned models, EchoPrompt captures contextual traces left by the generation process. Experiments on three benchmarks show that EchoPrompt achieves strong overall performance, including in the low-FPR regime, and robustness across domains, proxy models, text lengths, and adversarial attacks. These results highlight latent prompt dependency as an effective signal for zero-shot machine-generated text detection.

\section*{Ethics Statement}
EchoPrompt is intended to provide an auxiliary signal for identifying LLM-generated text. Like other automated detectors, it may produce false positives that incorrectly label human-written text as machine-generated, as well as false negatives that fail to identify generated content. Such errors can have greater consequences in high-stakes settings such as education, employment, publishing, and content moderation. We therefore do not recommend using EchoPrompt as definitive evidence of authorship or as the sole basis for consequential decisions.

\section*{Reproducibility Statement}
We provide the complete formulation and implementation details of EchoPrompt in Section~\ref{sec:method} and Section~\ref{sec:experiments}. The evaluated datasets, proxy models, baselines, and experimental settings are summarized in Section~\ref{sec:exp-setup}, with full details provided in Appendix~\ref{app:experimental-setup}.

\section*{AI Use Statement}
Generative AI tools were used to assist with language editing and improving the clarity of the manuscript. They were not used to generate experimental results, conduct data analysis, or make scientific conclusions. All AI-assisted text was reviewed and revised by the authors, who take full responsibility for the content of the paper.

\bibliography{custom}
\bibliographystyle{iclr2027_conference}

\appendix

\section{Empirical Study of Global Prefix Design}
\label{app:prefix-design}

We determine the global prefix $c_g$ through preliminary empirical studies under a fixed EchoPrompt setting. To avoid selecting the prefix on any evaluation benchmark, we use an independent subset of HC3~\citep{guo2023close}, consisting of 500 human-written answers and 500 ChatGPT-generated answers. We compare different prompt types and semantically related prompt families using Qwen2.5-3B / Qwen2.5-3B-Instruct, while keeping the remaining scoring configuration unchanged. The selected prefix is then fixed and used without further tuning across DetectRL, RealDet, RAID, and all evaluated proxy families.

Table~\ref{tab:cg-prompt-candidates} and Table~\ref{tab:cg-prompt-family-candidates} list the candidate prefixes. The resulting comparison shows the same overall pattern as our original preliminary study: semantically meaningful assistant-response contexts consistently outperform random or irrelevant prefixes, supporting the use of a generic response-oriented context rather than benchmark-specific prompt tuning.

\begin{table}[h]
\centering
\caption{Prefix-selection results on the independent HC3 development set.}
\label{tab:cg-prefix-independent}
\small
\setlength{\tabcolsep}{5pt}
\begin{tabular}{lcc}
\toprule
Prompt group & Mean AUROC & Std. \\
\midrule
System prompt & \textbf{0.8714} & 0.0287 \\
Role prefix & 0.8450 & 0.0161 \\
Random prefix & 0.7992 & 0.0198 \\
Irrelevant prefix & 0.7682 & 0.0275 \\
\midrule
Helpful-assistant family & \textbf{0.8595} & 0.0276 \\
Analytical-style family & 0.7710 & 0.0350 \\
Neutral-response family & 0.7558 & 0.0421 \\
\bottomrule
\end{tabular}
\end{table}

\begin{table*}[h]
\centering
\caption{Candidate prefixes in the prompt-type study.}
\label{tab:cg-prompt-candidates}
\small
\setlength{\tabcolsep}{5pt}
\renewcommand{\arraystretch}{1.08}
\begin{tabularx}{\textwidth}{llX}
\toprule
Prompt type & Prompt ID & Prefix text \\
\midrule
system\_prompt & \texttt{system\_helpful} & ``You are a helpful and reliable AI assistant. Answer clearly and avoid unsupported claims.'' \\
system\_prompt & \texttt{system\_precise} & ``You are a precise technical writing assistant. Use concise wording and keep factual consistency.'' \\
role\_prefix & \texttt{role\_journalist} & ``[Role: Investigative journalist writing for a global newspaper]'' \\
role\_prefix & \texttt{role\_professor} & ``[Role: University professor preparing lecture notes]'' \\
random\_prefix & \texttt{random\_noise\_a} & ``zxqv n11 4rpt k2lm w0a9 nnn pqr ttt'' \\
random\_prefix & \texttt{random\_noise\_b} & ``m5 m5 m5 uuu 9x 9x 9x qlmn zz y7'' \\
irrelevant\_prefix & \texttt{irrelevant\_cooking} & ``Recipe context: preheat oven to 180C, whisk eggs, fold flour, and bake for 25 minutes.'' \\
irrelevant\_prefix & \texttt{irrelevant\_travel} & ``Travel memo: train departs at platform 4, weather is rainy, carry a light umbrella and check hotel check-in time.'' \\
\bottomrule
\end{tabularx}
\end{table*}

\begin{table*}[h]
\centering
\caption{Candidate prefixes in the semantic-family study.}
\label{tab:cg-prompt-family-candidates}
\small
\setlength{\tabcolsep}{5pt}
\renewcommand{\arraystretch}{1.08}
\begin{tabularx}{\textwidth}{llX}
\toprule
Prompt family & Prompt ID & Prefix text \\
\midrule
helpful\_assistant & \texttt{family\_helpful\_v1} & ``You are a helpful AI assistant. Provide clear, factual, and coherent answers.'' \\
helpful\_assistant & \texttt{family\_helpful\_v2} & ``Act as a reliable assistant and answer with clear structure and factual consistency.'' \\
helpful\_assistant & \texttt{family\_helpful\_v3} & ``Serve as a trustworthy assistant: be concise, accurate, and logically organized.'' \\
analytical\_style & \texttt{family\_analytical\_v1} & ``Respond in an analytical style: state claims, then support each claim with evidence.'' \\
analytical\_style & \texttt{family\_analytical\_v2} & ``Answer with explicit reasoning steps, emphasizing precision and evidence-backed statements.'' \\
analytical\_style & \texttt{family\_analytical\_v3} & ``Use a structured analytical tone with concise argumentation and factual grounding.'' \\
neutral\_response & \texttt{family\_neutral\_v1} & ``Produce a neutral and well-formed response to the request.'' \\
neutral\_response & \texttt{family\_neutral\_v2} & ``Generate a balanced response that stays on topic and avoids unsupported assumptions.'' \\
neutral\_response & \texttt{family\_neutral\_v3} & ``Write a coherent, topic-focused answer with stable tone and objective wording.'' \\
\bottomrule
\end{tabularx}
\end{table*}

\section{Context-Conditioned Gain Under Assistant-Style Context}
\label{app:Context-Conditioned Gain}
Table~\ref{tab:context-gain-appendix} reports the base-model context-conditioned gains for the four proxy families used in the contextual restoration validation.

\begin{table*}[h]
\centering
\caption{Context-conditioned gain under an assistant-style prefix for the base models of four proxy families on 1,000 randomly sampled pairs from DetectRL Multi-Domain and DetectRL Multi-LLM. For each dataset and proxy family, the table reports the human-side and machine-side mean gain $g(X; c_g)$.}
\label{tab:context-gain-appendix}
\small
\setlength{\tabcolsep}{6pt}
\renewcommand{\arraystretch}{1.08}
\begin{tabular}{llcc}
\toprule
Model & Dataset & Human Gain & Machine Gain \\
\midrule
Qwen2.5-1.5B & DetectRL Multi-Domain & +0.0209 & +0.0486 \\
Qwen2.5-1.5B & DetectRL Multi-LLM & +0.0040 & +0.0461 \\
\midrule
Qwen2.5-3B & DetectRL Multi-Domain & +0.0056 & +0.0377 \\
Qwen2.5-3B & DetectRL Multi-LLM & -0.0062 & +0.0362 \\
\midrule
Llama-3.2-1B & DetectRL Multi-Domain & -0.0451 & +0.0261 \\
Llama-3.2-1B & DetectRL Multi-LLM & -0.0483 & +0.0257 \\
\midrule
Llama-3.2-3B & DetectRL Multi-Domain & -0.0381 & +0.0385 \\
Llama-3.2-3B & DetectRL Multi-LLM & -0.0417 & +0.0381 \\
\bottomrule
\end{tabular}
\end{table*}

\section{Detailed Experimental Setup}
\label{app:experimental-setup}

\noindent\textbf{Datasets.}
We evaluate EchoPrompt on three public detection benchmarks: DetectRL~\citep{wu2024detectrl}, RealDet~\citep{zhu2025reliably}, and RAID~\citep{dugan2024raid}. For DetectRL, we follow the standard test setting and use three deduplicated splits: Multi-Domain with 3,975 human/machine pairs, Multi-LLM with 3,991 pairs, and Multi-Attack with 4,967 pairs aggregated from all attack categories. To ensure fair and accurate auxiliary evaluation beyond DetectRL, we further use balanced subsets of 1,000 human and 1,000 machine samples from RealDet and RAID.

\noindent\textbf{Baselines.}
We compare EchoPrompt with representative recent training-based and training-free baselines. The training-based baselines include OpenAI-D~\citep{solaiman2019release}, BiScope~\citep{guo2024biscope}, and R-Detect~\citep{song2025deep}, while the training-free baselines include Likelihood~\citep{hashimoto2019unifying}, LogRank~\citep{gehrmann2019gltr}, Entropy~\citep{ippolito2020automatic}, Fast-DetectGPT~\citep{bao2023fast}, Binoculars~\citep{hans2024spotting}, LastDE++~\citep{xu2024training}, DNA-DetectLLM~\citep{zhu2025dna}, and IRM~\citep{liu2026zero}. These methods cover supervised detectors and major zero-shot detection strategies based on confidence, curvature, cross-model discrepancy, and alignment-aware signals.

\noindent\textbf{Models.}
To examine the effects of proxy family and model scale, we evaluate paired base/instruct models from Qwen2.5~\citep{qwen2_5}, Llama-3.2~\citep{meta2024llama32card}, Llama-3.1~\citep{meta2024llama31card}, Llama-3~\citep{meta2024llama3card,grattafiori2024llama}, and Falcon~\citep{almazrouei2023falcon}, including Qwen2.5-1.5B/3B, Llama-3.2-1B/3B, Llama-3.1-8B, Meta-Llama-3-8B, and Falcon-7B. For dual-model detectors, the base and instruct models are used as the observer/performer pair; for single-model detectors, the instruct model is used as the observer model.

\noindent\textbf{Metrics.}
We use \textbf{AUROC} and \textbf{Best-F1} as the primary evaluation metrics. AUROC measures the overall separability between human-written and machine-generated text and is independent of the decision threshold. Best-F1 denotes the maximum F1 score obtained by scanning the decision threshold on the evaluated scores, and is used to summarize score separability under thresholded detection. We additionally report TPR at 1\% FPR for low-false-positive-rate evaluation.

\noindent\textbf{Implementation details.}
For a fair comparison, all training-based detectors are trained on HC3~\citep{guo2023close}, which is disjoint from the evaluation benchmarks. Experiments are conducted on Tesla V100-PCIE-32GB GPUs with a maximum input length of 1024 tokens. The prompt setting follows Section~\ref{subsec:contextual_restoration}, and all other parameters use default values unless otherwise specified.

\section{Full Main Results Across Proxy Pairs}
\label{app:Full Main Results}
Table~\ref{tab:appendix-main-results-full} reports the complete main-results table, with the proxy-independent training-based block followed by all evaluated training-free proxy pairs.

\section{Full Per-Attack Results}
\label{app:Full Per-Attack Results}
Table~\ref{tab:appendix-attack-details} reports the complete per-attack AUROC and Best-F1 results, with the proxy-independent training-based block followed by all training-free proxy pairs.

\section{Low-FPR Evaluation}
\label{app:low-fpr}

To further compare EchoPrompt and IRM under a practical detection setting, we evaluate their true-positive rates when the false-positive rate is restricted to 1\%. In real-world applications, incorrectly identifying human-written text as machine-generated can have substantial consequences, making detection performance under a strict false-positive constraint particularly important. We therefore report TPR@1\%FPR using the Llama-3-8B proxy family on DetectRL~\citep{wu2024detectrl}, RealDet~\citep{zhu2025reliably}, and RAID~\citep{dugan2024raid}.

As shown in Table~\ref{tab:low-fpr}, EchoPrompt consistently achieves higher TPR than IRM across all five evaluation settings. The macro-average TPR@1\%FPR increases from \textbf{73.11\%} for IRM to \textbf{81.86\%} for EchoPrompt, corresponding to an improvement of \textbf{8.75} percentage points. This result shows that the benefit of restored assistant-response context is particularly pronounced in the low-FPR region, even when the overall AUROC of the two methods is already high.

\begin{table}[h]
\centering
\caption{TPR at 1\% FPR (\%) using the Llama-3-8B proxy family.}
\label{tab:low-fpr}
\small
\begin{tabular}{lcc}
\toprule
Dataset & EchoPrompt & IRM \\
\midrule
DetectRL Multi-Domain & \textbf{89.76} & 84.75 \\
DetectRL Multi-LLM & \textbf{90.55} & 83.54 \\
DetectRL Multi-Attack & \textbf{88.90} & 83.88 \\
RealDet & \textbf{69.40} & 51.20 \\
RAID & \textbf{70.70} & 62.20 \\
\midrule
Macro Average & \textbf{81.86} & 73.11 \\
\bottomrule
\end{tabular}
\end{table}

\section{Robustness Beyond Standard Attacks}
\label{app:targeted-robustness}

We further evaluate EchoPrompt under settings that move away from conventional assistant-style answers. For DetectRL~\citep{wu2024detectrl}, Yelp \texttt{direct\_prompt} asks the model to continue a review from a human-written first sentence; Few-shot provides human-written examples for imitation; ICO/SICO explicitly identifies human--AI lexical and syntactic differences and requests human-style generation; and Back-translation rewrites machine-generated text through an English--Chinese--English translation cycle. We additionally evaluate 1,000 balanced human--machine pairs from the non-adversarial split of RAID~\citep{dugan2024raid}, evenly sampled from GPT-2, GPT-3, Cohere, MPT, and Mistral outputs. All experiments use the Llama-3-8B base/instruction pair.

\begin{table}[h]
\centering
\caption{EchoPrompt under continuation, human-style, rewriting, and non-chat generation settings.}
\label{tab:targeted-robustness}
\small
\begin{tabular}{lcc}
\toprule
Setting & AUROC & Best-F1 \\
\midrule
Yelp continuation & 0.9998 & 0.9956 \\
Few-shot human writing & 0.9696 & 0.9262 \\
ICO/SICO human style & 0.9960 & 0.9764 \\
Back-translation & 0.9846 & 0.9531 \\
RAID non-chat/completion & 0.8250 & 0.7711 \\
\bottomrule
\end{tabular}
\end{table}

Continuation formatting, human-style prompting, and rewriting substantially alter the observable form of machine-generated text, while the underlying generation still passes through post-trained instruction-following models. EchoPrompt therefore retains a strong detection signal in these settings. In contrast, the lower performance on non-chat/completion outputs is consistent with the proposed mechanism: when generation is less dependent on instruction-tuned assistant behavior, the latent response-context dependency exploited by EchoPrompt can become weaker.

\section{Prompt-Component Ablation}
\label{app:prompt-ablation}
To analyze the contribution of each part of the generic prefix, we decompose it into three interpretable components. The \emph{role sentence} is ``You are a helpful, versatile, and intelligent AI assistant.'' The \emph{context clause A} is ``Below is the content you generated in response to a user's request,'' and \emph{context clause B} specifies that the continuation may take the form of ``a coherent continuation, topic-specific article, or detailed answer to a question.'' We conduct the ablation using two proxy families, Qwen2.5-3B and Llama-3.1-8B. For both DetectRL Multi-Domain and DetectRL Multi-LLM, we randomly sample 1,000 paired examples from each dataset.

\begingroup
\scriptsize
\setlength{\tabcolsep}{1.4pt}
\renewcommand{\arraystretch}{1.02}
\setlength{\LTleft}{\fill}
\setlength{\LTright}{\fill}

\begin{longtable}{lcccccccccccc}
\caption{Full main results (\%) across all proxy pairs and benchmarks. Bold and underlines mark the best and second-best results within each training-free proxy block.}
\label{tab:appendix-main-results-full}\\
\toprule
Detectors & \multicolumn{2}{c}{\shortstack{\textbf{DetectRL}\\\textbf{Multi-Domain}}} & \multicolumn{2}{c}{\shortstack{\textbf{DetectRL}\\\textbf{Multi-LLM}}} & \multicolumn{2}{c}{\shortstack{\textbf{DetectRL}\\\textbf{Multi-Attack}}} & \multicolumn{2}{c}{\textbf{RealDet}} & \multicolumn{2}{c}{\textbf{RAID}} & \multicolumn{2}{c}{\textbf{Avg}} \\
\cmidrule(lr){2-3}\cmidrule(lr){4-5}\cmidrule(lr){6-7}\cmidrule(lr){8-9}\cmidrule(lr){10-11}\cmidrule(lr){12-13}
& AUROC & Best-F1 & AUROC & Best-F1 & AUROC & Best-F1 & AUROC & Best-F1 & AUROC & Best-F1 & AUROC & Best-F1 \\
\midrule
\endfirsthead

\multicolumn{13}{c}{\tablename\ \thetable{} (continued)}\\
\toprule
Detectors & \multicolumn{2}{c}{\shortstack{\textbf{DetectRL}\\\textbf{Multi-Domain}}} & \multicolumn{2}{c}{\shortstack{\textbf{DetectRL}\\\textbf{Multi-LLM}}} & \multicolumn{2}{c}{\shortstack{\textbf{DetectRL}\\\textbf{Multi-Attack}}} & \multicolumn{2}{c}{\textbf{RealDet}} & \multicolumn{2}{c}{\textbf{RAID}} & \multicolumn{2}{c}{\textbf{Avg}} \\
\cmidrule(lr){2-3}\cmidrule(lr){4-5}\cmidrule(lr){6-7}\cmidrule(lr){8-9}\cmidrule(lr){10-11}\cmidrule(lr){12-13}
& AUROC & Best-F1 & AUROC & Best-F1 & AUROC & Best-F1 & AUROC & Best-F1 & AUROC & Best-F1 & AUROC & Best-F1 \\
\midrule
\endhead

\bottomrule
\endfoot

\bottomrule
\endlastfoot

\methodgrouprow{13}{Training-based Methods}
OpenAI-D & 82.64 & 76.00 & 82.58 & 75.61 & 83.91 & 78.02 & 90.06 & 77.15 & 73.03 & 67.19 & 82.44 & 74.80 \\
BiScope & 76.81 & 70.75 & 76.91 & 70.37 & 79.10 & 74.07 & 88.90 & 72.08 & 79.42 & 73.93 & 80.23 & 72.24 \\
R-Detect & 77.83 & 73.29 & 78.22 & 73.52 & 78.54 & 74.18 & 87.84 & 72.73 & 70.75 & 68.39 & 78.64 & 72.42 \\
\midrule
\methodgrouprow{13}{Training-free Methods}
\textit{Qwen2.5-1.5B family} & \multicolumn{12}{l}{} \\
Entropy & 73.33 & 67.80 & 73.33 & 67.99 & \underline{76.67} & \underline{74.20} & \underline{86.34} & 80.23 & 72.79 & 68.34 & 76.49 & 71.71 \\
Likelihood & 70.65 & 66.67 & 70.34 & 66.67 & 74.10 & 73.60 & 86.03 & \underline{81.75} & 75.79 & 72.29 & 75.38 & 72.20 \\
LogRank & 72.08 & 66.67 & 71.75 & 66.68 & 75.33 & 73.80 & \textbf{87.53} & \textbf{82.23} & 77.23 & 72.89 & \textbf{76.78} & \underline{72.45} \\
Fast-DetectGPT & 65.95 & 66.77 & 65.00 & 66.74 & 67.26 & 67.23 & 78.23 & 73.26 & \underline{77.52} & \underline{75.34} & 70.79 & 69.87 \\
Binoculars & 64.95 & 66.70 & 63.90 & 66.68 & 66.21 & 67.24 & 77.02 & 73.92 & 77.31 & \textbf{76.77} & 69.88 & 70.26 \\
LastDE++ & 53.43 & 66.68 & 52.16 & 66.68 & 68.79 & 68.22 & 74.23 & 69.39 & \textbf{78.65} & 74.57 & 65.45 & 69.11 \\
DNA-DetectLLM & 69.25 & 66.71 & 68.22 & 66.68 & 69.49 & 67.12 & 76.42 & 73.30 & 76.29 & 75.08 & 71.93 & 69.78 \\
IRM & \underline{75.23} & \underline{71.78} & \underline{74.56} & \underline{71.27} & 74.47 & 71.10 & 67.04 & 67.92 & 65.69 & 66.67 & 71.40 & 69.75 \\
EchoPrompt & \textbf{82.44} & \textbf{78.87} & \textbf{81.74} & \textbf{78.52} & \textbf{83.71} & \textbf{80.20} & 67.77 & 69.43 & 68.23 & 67.52 & \underline{76.78} & \textbf{74.91} \\
\midrule
\textit{Qwen2.5-3B family} & \multicolumn{12}{l}{} \\
Entropy & \underline{74.42} & \underline{69.17} & \underline{74.06} & \underline{68.55} & \textbf{77.80} & \underline{74.56} & \underline{84.79} & \underline{79.62} & 71.56 & 67.01 & \textbf{76.53} & \underline{71.78} \\
Likelihood & 70.38 & 66.68 & 69.93 & 66.67 & 73.52 & 73.36 & 83.86 & 79.56 & 75.27 & 71.79 & 74.59 & 71.61 \\
LogRank & 72.25 & 66.86 & 71.77 & 66.68 & 75.12 & 73.81 & \textbf{86.01} & \textbf{80.60} & \underline{76.69} & 72.26 & \underline{76.37} & \textbf{72.04} \\
Fast-DetectGPT & 67.84 & 67.38 & 67.47 & 67.38 & 69.67 & 69.19 & 74.67 & 69.94 & 76.30 & 74.07 & 71.19 & 69.59 \\
Binoculars & 66.29 & 66.94 & 66.04 & 66.98 & 69.22 & 68.83 & 72.96 & 69.76 & 76.11 & \textbf{74.76} & 70.12 & 69.45 \\
LastDE++ & 56.64 & 66.83 & 56.55 & 66.78 & 71.79 & 69.89 & 70.01 & 67.26 & \textbf{77.72} & 73.33 & 66.54 & 68.82 \\
DNA-DetectLLM & 73.10 & 67.40 & 72.65 & 67.13 & 74.07 & 69.81 & 73.18 & 69.47 & 75.15 & \underline{74.09} & 73.63 & 69.58 \\
IRM & 62.25 & 66.67 & 61.39 & 66.68 & 61.56 & 66.75 & 55.13 & 66.82 & 59.22 & 66.67 & 59.91 & 66.72 \\
EchoPrompt & \textbf{76.02} & \textbf{74.22} & \textbf{75.11} & \textbf{73.69} & \underline{77.38} & \textbf{74.95} & 63.24 & 67.55 & 66.94 & 67.50 & 71.74 & 71.58 \\
\midrule
\textit{Llama-3.2-1B family} & \multicolumn{12}{l}{} \\
Entropy & 75.34 & 69.88 & 74.87 & 69.06 & 78.48 & 76.74 & 86.88 & 80.77 & 76.15 & 72.15 & 78.34 & 73.72 \\
Likelihood & 79.71 & 73.91 & 79.36 & 73.51 & 82.61 & 78.19 & 87.74 & 83.85 & 77.89 & 74.21 & 81.46 & 76.73 \\
LogRank & 79.89 & 73.83 & 79.45 & 73.55 & 82.50 & 78.12 & 88.40 & 84.01 & 78.69 & 74.49 & 81.79 & 76.80 \\
Fast-DetectGPT & 90.76 & 83.25 & 90.43 & 82.85 & 91.66 & 84.64 & 89.43 & 85.14 & 83.69 & \underline{83.82} & 89.19 & 83.94 \\
Binoculars & 91.91 & 85.31 & 91.64 & 84.95 & 92.77 & 86.77 & 90.70 & \textbf{87.11} & 83.97 & \textbf{83.94} & 90.20 & 85.61 \\
LastDE++ & 79.99 & 72.48 & 79.33 & 71.62 & 82.27 & 77.07 & 83.81 & 77.07 & \underline{84.92} & 81.10 & 82.06 & 75.87 \\
DNA-DetectLLM & 93.07 & 87.01 & 92.76 & 86.15 & 93.73 & 88.01 & 91.05 & \underline{86.76} & 83.51 & 82.45 & 90.82 & 86.08 \\
IRM & \textbf{97.39} & \textbf{91.74} & \textbf{97.23} & \textbf{91.42} & \textbf{97.23} & \textbf{91.99} & \underline{91.32} & 82.71 & 84.65 & 79.53 & \textbf{93.56} & \textbf{87.48} \\
EchoPrompt & \underline{96.07} & \underline{89.71} & \underline{95.43} & \underline{88.74} & \underline{95.78} & \underline{90.54} & \textbf{92.48} & 86.62 & \textbf{86.98} & 81.73 & \underline{93.35} & \underline{87.47} \\
\midrule
\textit{Llama-3.2-3B family} & \multicolumn{12}{l}{} \\
Entropy & 74.93 & 69.31 & 74.36 & 67.81 & 77.25 & 75.44 & 83.97 & 79.09 & 74.61 & 70.81 & 77.03 & 72.49 \\
Likelihood & 79.55 & 73.88 & 79.20 & 73.22 & 82.18 & 77.29 & 86.84 & 82.51 & 76.78 & 73.30 & 80.91 & 76.04 \\
LogRank & 79.84 & 73.76 & 79.41 & 73.06 & 82.24 & 77.38 & 87.65 & 82.60 & 77.82 & 73.92 & 81.39 & 76.15 \\
Fast-DetectGPT & 90.03 & 82.17 & 89.75 & 81.88 & 90.71 & 83.51 & 89.47 & 83.31 & 84.02 & \underline{83.09} & 88.80 & 82.79 \\
Binoculars & 91.19 & 84.42 & 90.88 & 83.91 & 91.73 & 85.43 & \underline{90.66} & \textbf{85.77} & 84.36 & \textbf{83.20} & 89.76 & 84.55 \\
LastDE++ & 79.89 & 72.61 & 79.52 & 71.95 & 81.91 & 76.58 & 83.39 & 75.87 & \textbf{85.94} & 80.94 & 82.13 & 75.59 \\
DNA-DetectLLM & 92.55 & 85.86 & 92.15 & 85.55 & 92.82 & 86.93 & 90.50 & \underline{85.34} & 83.27 & 81.04 & 90.26 & 84.94 \\
IRM & \textbf{97.27} & \textbf{91.80} & \textbf{97.23} & \textbf{91.49} & \textbf{97.25} & \underline{91.74} & 88.94 & 81.11 & 82.75 & 76.96 & \underline{92.69} & \underline{86.62} \\
EchoPrompt & \underline{97.01} & \underline{91.24} & \underline{96.55} & \underline{90.93} & \underline{96.87} & \textbf{91.91} & \textbf{90.84} & 85.33 & \underline{84.41} & 78.36 & \textbf{93.14} & \textbf{87.55} \\
\midrule
\textit{Falcon-7B family} & \multicolumn{12}{l}{} \\
Entropy & 66.29 & 66.68 & 66.10 & 66.69 & 72.45 & 72.89 & 87.08 & 81.23 & 72.28 & 68.06 & 72.84 & 71.11 \\
Likelihood & 68.09 & 66.68 & 67.88 & 66.69 & 72.39 & 72.44 & 86.25 & 81.47 & 75.16 & 72.11 & 73.95 & 71.88 \\
LogRank & 68.63 & 66.68 & 68.29 & 66.69 & 72.20 & 72.66 & 87.35 & 81.77 & 76.28 & 72.59 & 74.55 & 72.08 \\
Fast-DetectGPT & 82.80 & 76.15 & 82.33 & 75.92 & 84.04 & 77.30 & 90.81 & 86.98 & 84.84 & \underline{86.53} & 84.96 & 80.58 \\
Binoculars & 83.87 & 78.28 & 83.29 & 78.25 & 85.13 & 79.42 & \underline{91.40} & \underline{88.31} & 84.95 & 86.40 & 85.73 & \underline{82.13} \\
LastDE++ & 75.81 & 69.71 & 75.38 & 69.06 & 78.67 & 74.97 & 90.33 & 84.12 & \textbf{89.05} & 86.28 & 81.85 & 76.83 \\
DNA-DetectLLM & \textbf{89.27} & \textbf{83.01} & \textbf{88.87} & \textbf{82.45} & \textbf{90.02} & \textbf{83.81} & \textbf{92.45} & \textbf{88.64} & \underline{86.48} & \textbf{86.98} & \textbf{89.42} & \textbf{84.98} \\
IRM & 82.83 & 76.27 & 82.62 & 75.85 & 84.54 & 78.02 & 81.59 & 74.32 & 78.31 & 72.51 & 81.98 & 75.39 \\
EchoPrompt & \underline{87.82} & \underline{81.62} & \underline{87.36} & \underline{80.96} & \underline{88.64} & \underline{82.68} & 86.89 & 83.52 & 83.84 & 79.80 & \underline{86.91} & 81.72 \\
\midrule
\textit{Llama-3.1-8B family} & \multicolumn{12}{l}{} \\
Entropy & 68.45 & 66.68 & 68.01 & 66.68 & 71.32 & 72.60 & 80.16 & 74.53 & 69.74 & 66.69 & 71.54 & 69.44 \\
Likelihood & 74.88 & 69.17 & 74.59 & 68.19 & 77.51 & 74.11 & 83.51 & 79.19 & 73.91 & 70.33 & 76.88 & 72.20 \\
LogRank & 74.94 & 69.05 & 74.58 & 68.21 & 77.40 & 74.02 & 84.26 & 79.37 & 74.89 & 70.96 & 77.22 & 72.32 \\
Fast-DetectGPT & 85.19 & 77.14 & 84.99 & 76.97 & 86.71 & 79.82 & 86.59 & 79.32 & 82.69 & \textbf{81.71} & 85.24 & 78.99 \\
Binoculars & 85.93 & 78.39 & 85.81 & 78.10 & 87.38 & 80.88 & 87.33 & \underline{81.81} & 82.91 & 81.35 & 85.87 & 80.11 \\
LastDE++ & 77.22 & 70.15 & 76.77 & 69.95 & 80.02 & 76.24 & 83.16 & 75.95 & \textbf{86.50} & \underline{81.44} & 80.73 & 74.75 \\
DNA-DetectLLM & 87.16 & 79.92 & 86.94 & 79.56 & 88.56 & 81.46 & \underline{87.41} & 81.64 & 81.01 & 79.12 & \underline{86.22} & \underline{80.34} \\
IRM & \underline{88.83} & \underline{82.53} & \underline{88.96} & \underline{82.35} & \underline{88.83} & \underline{82.87} & 82.23 & 75.90 & 79.37 & 72.10 & 85.65 & 79.15 \\
EchoPrompt & \textbf{95.10} & \textbf{89.13} & \textbf{95.37} & \textbf{89.50} & \textbf{95.61} & \textbf{89.90} & \textbf{88.88} & \textbf{82.67} & \underline{85.21} & 78.10 & \textbf{92.03} & \textbf{85.86} \\
\midrule
\textit{Llama-3-8B family} & \multicolumn{12}{l}{} \\
Entropy & 64.31 & 66.71 & 64.14 & 66.68 & 67.64 & 71.06 & 77.08 & 73.04 & 67.73 & 66.69 & 68.18 & 68.84 \\
Likelihood & 79.54 & 73.57 & 79.24 & 72.75 & 81.71 & 76.12 & 85.53 & 81.18 & 75.89 & 73.11 & 80.38 & 75.34 \\
LogRank & 76.83 & 70.70 & 76.39 & 69.63 & 79.17 & 74.63 & 85.34 & 80.81 & 75.80 & 72.49 & 78.71 & 73.65 \\
Fast-DetectGPT & 91.40 & 84.02 & 91.45 & 83.61 & 92.63 & 85.83 & 88.90 & 82.96 & 83.68 & 83.01 & 89.61 & 83.89 \\
Binoculars & 91.93 & 84.98 & 92.01 & 84.57 & 93.10 & 86.61 & 89.58 & \underline{84.46} & 83.73 & 82.49 & 90.07 & 84.62 \\
LastDE++ & 84.26 & 76.72 & 84.26 & 76.74 & 86.69 & 80.43 & 89.10 & 81.74 & \underline{87.81} & \underline{83.94} & 86.42 & 79.91 \\
DNA-DetectLLM & 90.41 & 83.28 & 90.55 & 82.99 & 91.86 & 85.05 & 87.75 & 82.42 & 80.94 & 80.55 & 88.30 & 82.86 \\
IRM & \underline{98.47} & \underline{94.06} & \underline{98.50} & \underline{93.89} & \underline{98.36} & \underline{93.92} & \underline{91.51} & 83.43 & 87.52 & 81.42 & \underline{94.87} & \underline{89.34} \\
EchoPrompt & \textbf{98.82} & \textbf{95.26} & \textbf{98.62} & \textbf{95.25} & \textbf{98.60} & \textbf{95.28} & \textbf{92.33} & \textbf{88.79} & \textbf{89.45} & \textbf{85.32} & \textbf{95.56} & \textbf{91.98} \\
\end{longtable}
\endgroup

\begingroup
\scriptsize
\setlength{\tabcolsep}{2.4pt}
\renewcommand{\arraystretch}{1.02}
\setlength{\LTleft}{\fill}
\setlength{\LTright}{\fill}

\begin{longtable}{lcccccccccc}
\caption{Per-attack results (\%) on DetectRL attack subsets. Bold and underlines mark the best and second-best results within each training-free proxy block.}
\label{tab:appendix-attack-details}\\
\toprule
Detectors & \multicolumn{2}{c}{\textbf{Direct Prompt}} & \multicolumn{2}{c}{\textbf{Prompt Attacks}} & \multicolumn{2}{c}{\textbf{Paraphrase}} & \multicolumn{2}{c}{\textbf{Perturbation}} & \multicolumn{2}{c}{\textbf{Data Mixing}} \\
\cmidrule(lr){2-3}\cmidrule(lr){4-5}\cmidrule(lr){6-7}\cmidrule(lr){8-9}\cmidrule(lr){10-11}
& AUROC & Best-F1 & AUROC & Best-F1 & AUROC & Best-F1 & AUROC & Best-F1 & AUROC & Best-F1 \\
\midrule
\endfirsthead

\multicolumn{11}{c}{\tablename\ \thetable{} (continued)}\\
\toprule
Detectors & \multicolumn{2}{c}{\textbf{Direct Prompt}} & \multicolumn{2}{c}{\textbf{Prompt Attacks}} & \multicolumn{2}{c}{\textbf{Paraphrase}} & \multicolumn{2}{c}{\textbf{Perturbation}} & \multicolumn{2}{c}{\textbf{Data Mixing}} \\
\cmidrule(lr){2-3}\cmidrule(lr){4-5}\cmidrule(lr){6-7}\cmidrule(lr){8-9}\cmidrule(lr){10-11}
& AUROC & Best-F1 & AUROC & Best-F1 & AUROC & Best-F1 & AUROC & Best-F1 & AUROC & Best-F1 \\
\midrule
\endhead

\bottomrule
\endfoot

\bottomrule
\endlastfoot

\methodgrouprow{11}{Training-based Methods}
OpenAI-D & 93.12 & 83.65 & 88.08 & 78.71 & 87.48 & 79.12 & 71.00 & 72.01 & 79.85 & 76.62 \\
BiScope & 93.61 & 86.43 & 91.02 & 83.06 & 71.68 & 66.67 & 67.96 & 67.54 & 71.24 & 66.67 \\
R-Detect & 90.09 & 80.34 & 85.31 & 76.96 & 86.76 & 78.16 & 62.09 & 68.76 & 68.46 & 66.69 \\
\midrule
\methodgrouprow{11}{Training-free Methods}
\textit{Qwen2.5-1.5B family} & \multicolumn{10}{l}{} \\
Entropy & \underline{93.40} & \underline{86.43} & \textbf{91.08} & \textbf{83.80} & 67.44 & 66.69 & 61.51 & 67.30 & 69.92 & 66.76 \\
Likelihood & 92.82 & 86.03 & 88.52 & 81.96 & 69.79 & 66.67 & 51.93 & 66.67 & 67.41 & 66.67 \\
LogRank & \textbf{93.50} & \textbf{86.62} & \underline{89.16} & \underline{82.37} & 70.41 & 66.67 & 54.28 & 66.67 & 69.27 & 66.67 \\
Fast-DetectGPT & 75.37 & 69.13 & 70.43 & 66.73 & 70.46 & 66.84 & 54.11 & 66.78 & 65.94 & 66.67 \\
Binoculars & 74.30 & 69.42 & 69.54 & 66.69 & 69.55 & 66.73 & 52.33 & 66.69 & 65.35 & 66.67 \\
LastDE++ & 74.29 & 68.40 & 69.41 & 66.71 & 66.55 & 66.84 & \underline{78.53} & 72.48 & 55.19 & 66.67 \\
DNA-DetectLLM & 73.61 & 68.44 & 68.94 & 66.69 & 71.98 & 67.05 & 63.45 & 66.75 & 69.46 & 66.67 \\
IRM & 71.07 & 68.94 & 69.69 & 68.13 & \underline{76.20} & \underline{72.49} & 77.60 & \underline{73.90} & \underline{77.80} & \underline{72.05} \\
EchoPrompt & 81.19 & 78.60 & 81.60 & 78.75 & \textbf{80.34} & \textbf{78.55} & \textbf{84.91} & \textbf{82.10} & \textbf{90.49} & \textbf{83.00} \\
\midrule
\textit{Qwen2.5-3B family} & \multicolumn{10}{l}{} \\
Entropy & \underline{92.78} & \underline{85.94} & \textbf{90.17} & \textbf{82.98} & 65.42 & 66.69 & 67.05 & 68.87 & 73.60 & 68.31 \\
Likelihood & 92.32 & 84.86 & 88.72 & 81.96 & 68.30 & 66.67 & 51.95 & 66.67 & 66.31 & 66.67 \\
LogRank & \textbf{93.47} & \textbf{86.41} & \underline{89.38} & \underline{82.63} & 69.08 & 66.67 & 55.10 & 66.69 & 68.55 & 66.67 \\
Fast-DetectGPT & 81.22 & 73.68 & 77.03 & 70.63 & 71.87 & 67.71 & 50.43 & 67.25 & 67.79 & 66.67 \\
Binoculars & 80.50 & 73.47 & 76.14 & 70.09 & 70.74 & 67.25 & 52.29 & 66.67 & 66.41 & 66.67 \\
LastDE++ & 80.25 & 72.96 & 75.83 & 69.75 & 69.10 & 67.75 & \textbf{76.34} & \textbf{72.30} & 57.43 & 66.67 \\
DNA-DetectLLM & 81.61 & 74.63 & 77.06 & 70.25 & \underline{74.42} & \underline{68.55} & 62.65 & 67.04 & \underline{74.60} & \underline{68.60} \\
IRM & 63.80 & 66.67 & 61.72 & 66.94 & 67.99 & 66.83 & 53.16 & 66.67 & 61.12 & 66.67 \\
EchoPrompt & 79.24 & 76.85 & 79.68 & 77.44 & \textbf{75.09} & \textbf{73.71} & \underline{70.31} & \underline{70.96} & \textbf{82.55} & \textbf{75.79} \\
\midrule
\textit{Llama-3.2-1B family} & \multicolumn{10}{l}{} \\
Entropy & 97.20 & 92.32 & 95.04 & 89.50 & 71.88 & 68.55 & 57.37 & 66.69 & 70.88 & 66.67 \\
Likelihood & 96.99 & 91.69 & 93.57 & 87.34 & 76.10 & 71.69 & 67.49 & 66.69 & 78.90 & 73.56 \\
LogRank & 97.02 & 92.03 & 93.21 & 87.49 & 76.43 & 71.84 & 67.16 & 66.69 & 78.69 & 72.53 \\
Fast-DetectGPT & 98.26 & 93.12 & 94.92 & 88.52 & 88.86 & 81.09 & 86.82 & 79.06 & 89.44 & 81.43 \\
Binoculars & \underline{99.07} & \underline{95.67} & 96.13 & 91.17 & 90.21 & 83.33 & 88.18 & 81.06 & 90.24 & 82.59 \\
LastDE++ & 94.88 & 87.92 & 90.50 & 83.52 & 82.88 & 75.12 & 63.89 & 66.69 & 79.19 & 72.07 \\
DNA-DetectLLM & \textbf{99.51} & \textbf{96.63} & \underline{97.38} & \textbf{93.14} & 90.14 & 83.63 & 90.95 & 83.86 & 90.65 & 82.77 \\
IRM & 98.94 & 95.48 & \textbf{97.62} & \underline{92.83} & \textbf{98.00} & \textbf{92.51} & \textbf{95.65} & \textbf{88.62} & \underline{95.95} & \underline{90.48} \\
EchoPrompt & 97.52 & 94.70 & 96.49 & 91.90 & \underline{96.96} & \underline{91.80} & \underline{95.04} & \underline{87.78} & \textbf{96.47} & \textbf{91.17} \\
\midrule
\textit{Llama-3.2-3B family} & \multicolumn{10}{l}{} \\
Entropy & 97.26 & 92.56 & 95.05 & 89.50 & 72.03 & 68.57 & 56.84 & 66.69 & 71.32 & 66.67 \\
Likelihood & 97.14 & 91.85 & 93.65 & 87.87 & 76.28 & 71.50 & 67.50 & 66.68 & 79.04 & 72.69 \\
LogRank & 97.16 & 91.88 & 93.29 & 87.39 & 76.50 & 71.34 & 67.11 & 66.69 & 78.78 & 72.55 \\
Fast-DetectGPT & 98.27 & 93.43 & 94.95 & 88.75 & 88.57 & 80.93 & 85.41 & 77.86 & 88.84 & 80.68 \\
Binoculars & \underline{98.77} & \underline{94.61} & 95.80 & 90.40 & 89.41 & 81.91 & 87.06 & 79.11 & 89.83 & 81.61 \\
LastDE++ & 95.09 & 88.12 & 90.37 & 83.34 & 82.53 & 74.95 & 64.90 & 66.67 & 79.76 & 72.86 \\
DNA-DetectLLM & 98.22 & 93.67 & \underline{96.79} & \underline{91.72} & \underline{91.61} & \underline{84.72} & \underline{93.24} & \underline{85.76} & \underline{92.02} & \underline{84.15} \\
IRM & \textbf{99.22} & \textbf{95.73} & \textbf{97.76} & \textbf{93.09} & \textbf{98.34} & \textbf{93.41} & \textbf{98.24} & \textbf{93.78} & \textbf{96.57} & \textbf{90.81} \\
EchoPrompt & 98.58 & 94.79 & 96.61 & 91.30 & 96.08 & 89.79 & 96.38 & 90.47 & 94.43 & 87.87 \\
\midrule
\textit{Falcon-7B family} & \multicolumn{10}{l}{} \\
Entropy & 87.58 & 80.60 & 85.69 & 79.15 & 69.85 & 67.31 & 71.67 & 68.09 & 77.45 & 71.81 \\
Likelihood & 84.99 & 78.33 & 83.42 & 76.72 & 71.77 & 67.76 & 71.78 & 67.65 & 79.98 & 72.68 \\
LogRank & 84.79 & 77.91 & 83.31 & 76.58 & 71.67 & 67.70 & 72.26 & 67.78 & 80.23 & 72.82 \\
Fast-DetectGPT & 95.34 & 89.88 & 92.91 & 86.97 & 81.95 & 74.31 & 80.64 & 74.26 & 89.34 & 81.27 \\
Binoculars & 95.87 & 90.89 & 93.35 & 87.73 & 82.40 & 75.07 & 81.27 & 74.98 & 89.76 & 81.80 \\
LastDE++ & 91.58 & 84.43 & 88.23 & 80.39 & 76.57 & 69.21 & 72.34 & 67.51 & 83.35 & 76.18 \\
DNA-DetectLLM & \underline{96.53} & \underline{91.73} & \underline{94.40} & \underline{88.89} & \underline{83.66} & \underline{75.94} & \underline{84.27} & \underline{76.84} & \underline{91.21} & \underline{82.99} \\
IRM & 85.94 & 78.80 & 84.26 & 77.13 & 72.74 & 68.26 & 72.66 & 68.03 & 80.53 & 73.02 \\
EchoPrompt & \textbf{96.65} & \textbf{92.28} & \textbf{94.51} & \textbf{89.26} & \textbf{85.87} & \textbf{79.64} & \textbf{88.80} & \textbf{83.19} & \textbf{92.55} & \textbf{85.17} \\
\midrule
\textit{Llama-3.1-8B family} & \multicolumn{10}{l}{} \\
Entropy & 91.20 & 82.72 & 88.56 & 80.04 & 59.22 & 66.67 & 55.86 & 66.91 & 61.74 & 66.67 \\
Likelihood & 94.11 & 87.19 & 90.70 & 83.17 & 68.32 & 66.67 & 63.88 & 66.85 & 70.53 & 66.67 \\
LogRank & 93.97 & 87.01 & 90.16 & 82.77 & 68.28 & 66.67 & 64.24 & 66.94 & 70.37 & 66.69 \\
Fast-DetectGPT & 94.45 & 87.96 & 91.96 & 85.50 & 84.39 & 76.87 & 77.37 & 71.35 & 85.36 & 77.41 \\
Binoculars & 95.14 & 89.61 & 92.71 & 87.14 & 84.93 & 77.59 & 78.24 & 72.47 & 85.91 & 77.60 \\
LastDE++ & 92.48 & 85.41 & 89.50 & 82.15 & 82.18 & 75.33 & 58.31 & 66.69 & 77.60 & 71.62 \\
DNA-DetectLLM & \underline{95.76} & \underline{89.74} & \underline{93.52} & \underline{87.59} & 82.49 & 74.73 & 83.67 & 76.46 & 87.35 & 78.78 \\
IRM & 85.81 & 80.42 & 86.46 & 80.45 & \underline{90.78} & \underline{84.54} & \underline{90.47} & \underline{84.30} & \underline{90.62} & \underline{84.65} \\
EchoPrompt & \textbf{96.01} & \textbf{90.56} & \textbf{95.21} & \textbf{89.06} & \textbf{94.85} & \textbf{88.89} & \textbf{97.32} & \textbf{92.52} & \textbf{94.66} & \textbf{88.46} \\
\midrule
\textit{Llama-3-8B family} & \multicolumn{10}{l}{} \\
Entropy & 86.37 & 78.37 & 84.39 & 76.73 & 56.00 & 66.67 & 52.02 & 66.87 & 59.43 & 66.67 \\
Likelihood & 95.64 & 89.39 & 92.33 & 85.35 & 73.31 & 68.66 & 71.77 & 67.60 & 75.48 & 69.59 \\
LogRank & 94.86 & 88.26 & 91.09 & 83.94 & 70.75 & 66.67 & 66.79 & 67.20 & 72.36 & 67.08 \\
Fast-DetectGPT & 97.45 & 92.52 & 95.32 & 89.43 & 90.84 & 83.68 & 88.22 & 80.16 & 91.31 & 83.36 \\
Binoculars & 97.89 & 93.30 & 95.91 & 90.88 & 91.15 & 84.11 & 88.81 & 80.78 & 91.72 & 83.96 \\
LastDE++ & 96.72 & 90.50 & 93.91 & 87.95 & 88.59 & 81.03 & 70.35 & 66.75 & 83.87 & 75.92 \\
DNA-DetectLLM & 97.40 & 92.35 & 95.84 & 90.40 & 87.24 & 79.60 & 88.36 & 79.96 & 90.48 & 82.93 \\
IRM & \underline{98.70} & \underline{94.59} & \underline{97.73} & \underline{92.76} & \textbf{99.15} & \textbf{95.63} & \underline{98.90} & \underline{94.84} & \underline{97.31} & \underline{91.76} \\
EchoPrompt & \textbf{99.62} & \textbf{97.33} & \textbf{98.13} & \textbf{94.81} & \underline{98.28} & \underline{94.53} & \textbf{99.24} & \textbf{97.04} & \textbf{97.74} & \textbf{92.70} \\
\end{longtable}
\endgroup

\end{document}